\documentclass[lettersize,journal]{IEEEtran}
\usepackage{amsmath,amsfonts}
\usepackage{algorithmic}
\usepackage{algorithm}
\usepackage{array}
\usepackage{multirow}
\usepackage[caption=false,font=normalsize,labelfont=rm,textfont=rm]{subfig}
\usepackage{textcomp}
\usepackage{stfloats}
\usepackage{url}
\usepackage{verbatim}
\usepackage{graphicx}
\usepackage{cite}

\usepackage[table,xcdraw]{xcolor}
\usepackage{booktabs}

\usepackage{placeins}
\usepackage[hidelinks]{hyperref}

\begin{document}

\title{EECTracker: Swarm Motion Prior-Guided Feature Compensation for Airborne Optical UAV Swarm Tracking}

\markboth{IEEE Transactions on Geoscience and Remote Sensing}{}

\author{
	Zhaochen Chu, Tao Song, Ren Jin*, Mingdong Jia, Defu Lin
	\thanks{Zhaochen Chu, Tao Song, Ren Jin*, Mingdong Jia, Defu Lin are with the China-UAE Belt and Road Joint Laboratory on Intelligent Unmanned Systems, School of Aerospace Engineering, Beijing Institute of Technology, Beijing, 100081, China, (e-mail: \href{2315228186@qq.com}{2315228186@qq.com}; \href{6120160130@bit.edu.cn}{6120160130@bit.edu.cn}; \href{renjin@bit.edu.cn}{renjin@bit.edu.cn}; \href{jmd@bit.edu.cn}{jmd@bit.edu.cn}; \href{lindf@bit.edu.cn}{lindf@bit.edu.cn}). \itshape (Corresponding author: Ren Jin)}
	\thanks{
		This work has been submitted to the IEEE for possible publication. Copyright may be transferred without notice, after which this version may no longer be accessible.
}}
\maketitle

\begin{abstract}
	Airborne optical tracking of uncrewed aerial vehicle (UAV) swarms is challenging due to extremely small target scales, rapid viewpoint changes, and cluttered backgrounds, which can weaken target feature responses and lead to intermittent or temporarily missing detector responses. Existing multi-object tracking methods generally depend on reliable target-specific detector responses to maintain target states and identities across frames. When such responses become unreliable, target states cannot be reliably updated and cross-frame association cues become ambiguous, resulting in fragmented trajectories and identity switches. To address this problem, we propose EECTracker, a swarm-motion-prior-guided joint detection-and-tracking framework for airborne optical UAV swarm tracking. EECTracker constructs a probabilistic swarm motion prior from reliable historical tracklets to capture the shared short-term image-plane motion tendency of the swarm and its uncertainty, providing spatial guidance for cross-frame feature compensation. Building on this prior, we introduce Energy--Entropy Consistency Activation (EEC Activation) to evaluate motion-prior-conditioned feature consistency using feature residual energy and local residual entropy. The resulting Local EEC score guides pixel-level feature compensation by enhancing motion-prior-consistent feature responses in potential target regions while suppressing inconsistent background responses. Experiments on AIRMOT and UAVSwarm show that EECTracker achieves superior overall tracking performance compared with state-of-the-art methods. Compared with the strongest competing method SCT-MOT, EECTracker improves MOTA/IDF1 by 3.89/1.79 percentage points on AIRMOT and by 2.81/1.74 percentage points on UAVSwarm, while maintaining online inference speed.
\end{abstract}

\begin{IEEEkeywords}
UAV swarm tracking, multiple-object tracking, motion prior, feature compensation.
\end{IEEEkeywords}

\section{Introduction}
\IEEEPARstart{A}{irborne} optical tracking of uncrewed aerial vehicle (UAV) swarms is an important perception task for persistent airspace monitoring, cooperative UAV operations, and low-altitude situational awareness\cite{Anti-UAT,antiUAV410}. The task requires continuously localizing multiple UAV targets and maintaining their identities over time. Compared with ground-based optical sensing, airborne optical platforms provide more flexible viewpoints and broader observation coverage~\cite{Li2024TAV}, but are also challenged by extremely small target scales, rapid viewpoint changes, and complex background interference~\cite{yuan2023cfinet}. These factors can weaken target feature responses, making target observations intermittent or temporarily missing across consecutive frames and thereby increasing the difficulty of target localization and identity association over time.

Existing multi-object tracking (MOT) methods can be broadly categorized into tracking-by-detection (TBD) and joint detection and tracking (JDT) frameworks~\cite{fairmot,bytetrack}. TBD methods first localize candidate targets and then associate them across frames using appearance and motion cues, whereas JDT methods share feature representations for detection and identity association within a unified framework. More recently, transformer-based trackers~\cite{zeng2022motr,motrv3} have further unified detection and tracking in end-to-end architectures, where persistent track queries interact with current-frame features through cross-frame attention to maintain target states over time. In addition, closed-loop detection-tracking methods~\cite{shen2023interactivelyIMANet,chu2026sctmot} have strengthened the interaction between detection and tracking by feeding historical detections, motion predictions, or propagated features back to the current frame, thereby guiding current-frame detection and cross-frame feature propagation through detection-derived spatial references. Nevertheless, despite their architectural differences, the effectiveness of existing MOT methods generally depends on sufficiently reliable observations and historical states maintained for individual targets to propagate target states across frames, provide spatial cues for temporal feature interaction, and maintain identity associations.

Such dependence on sufficiently reliable target observations and historical states becomes particularly limiting in airborne optical UAV swarm tracking, where extremely small target scales, rapid viewpoint changes, and complex background interference can weaken target feature responses and make target observations intermittent or temporarily missing across adjacent frames~\cite{11027675TAES,chen2022local}. When target observations or historical states become unreliable, temporal feature propagation and cross-frame association are weakened, leading to trajectory fragmentation and identity switches. However, UAVs within a swarm often exhibit shared short-term motion tendencies in the image plane rather than moving independently. Even when some UAVs are weakly observed, the reliable tracklets that remain available can still provide shared motion information, which can be aggregated into a swarm-level motion prior to indicate potential UAV regions consistent with the inferred short-term motion tendency. Historical feature information can then be guided toward these regions to supplement weakened current-frame target responses. We refer to this swarm-motion-guided supplementation of weakened target feature responses as feature compensation.

To exploit this swarm-level motion information, we develop a swarm-motion-prior-guided feature compensation strategy for airborne optical UAV swarm tracking. Rather than deriving target-specific spatial guidance for feature compensation from reliable observations and historical states of every UAV, the strategy models individual image-plane motions using available reliable historical tracklets and aggregates them into a probabilistic swarm motion prior. The resulting prior captures the shared short-term motion tendency and uncertainty of the swarm, thereby characterizing statistically reachable image-plane regions of swarm targets in the current frame. It further provides dense pixel-level guidance across adjacent-frame feature maps to direct historical feature transfer and compensate the feature responses in potential target regions.

With the probabilistic swarm motion prior, the key challenge becomes how to identify reliable target-associated historical feature responses for current-frame feature compensation. Existing temporal feature alignment and enhancement methods commonly align or aggregate historical features using detection-derived locations or motion-predicted positions; therefore, their effectiveness degrades when these spatial references become unavailable under intermittent or temporarily missing target observations~\cite{wang2025omnitracker}. To address this limitation, we design Energy--Entropy Consistency Activation (EEC Activation), a pixel-level feature compensation mechanism built on motion-prior-conditioned feature consistency. Under the swarm motion prior, displacement-conditioned historical feature responses are compared with current-frame responses to form feature residuals. Feature residual energy measures the expected magnitude of these residuals, while local residual entropy characterizes their spatial dispersion. These two complementary statistics are then combined to construct the Local EEC score, which represents the local feature consistency between prior-guided historical responses and current-frame responses. This score is further converted into a channel-wise soft consistency mask to selectively weight feature-consistent historical responses during feature fusion. In this way, EEC Activation enhances feature-consistent responses in potential UAV target regions while suppressing feature-inconsistent background responses.

Building upon the proposed feature compensation strategy and EEC Activation, we further develop EECTracker, a swarm-motion-prior-guided joint detection and tracking framework for airborne optical UAV swarm tracking. At each frame, reliable historical tracklets are used to construct a probabilistic swarm motion prior, which guides feature compensation across adjacent-frame feature maps, while EEC Activation selectively controls historical feature fusion. The compensated feature representation is then used for target detection and association, and the updated tracklets are fed back to construct the motion prior for subsequent frames. Within this closed loop, historical tracklets provide swarm-level motion guidance for feature compensation, whereas the compensated features in turn support more reliable detection and association, thereby improving detection reliability and trajectory continuity under weakened target feature responses and intermittent target observations.

The main contributions of this work are summarized as follows:
\begin{enumerate}
	
	\item[1.] We develop a swarm-motion-prior-guided feature compensation strategy that constructs a probabilistic swarm motion prior from available reliable historical tracklets. By capturing the shared short-term motion tendency and uncertainty of the swarm, the prior provides dense pixel-level guidance for feature compensation without requiring reliable target-specific observations for every UAV.
	
	\item[2.] We propose Energy--Entropy Consistency Activation (EEC Activation), a pixel-level feature compensation mechanism that adaptively weights historical responses over potential target regions according to motion-prior-conditioned feature consistency. It combines feature residual energy and local residual entropy into a Local EEC score and converts the score into a channel-wise soft consistency mask for selective feature fusion.
	
	\item[3.] Building upon the above two components, we further develop EECTracker, a swarm-motion-prior-guided joint detection and tracking framework. Extensive experiments on UAV swarm tracking benchmarks demonstrate improved overall tracking performance and detection robustness compared with state-of-the-art methods.
	
\end{enumerate}

\begin{figure}[htbp]
	\centering
	\includegraphics[width=0.5\textwidth]{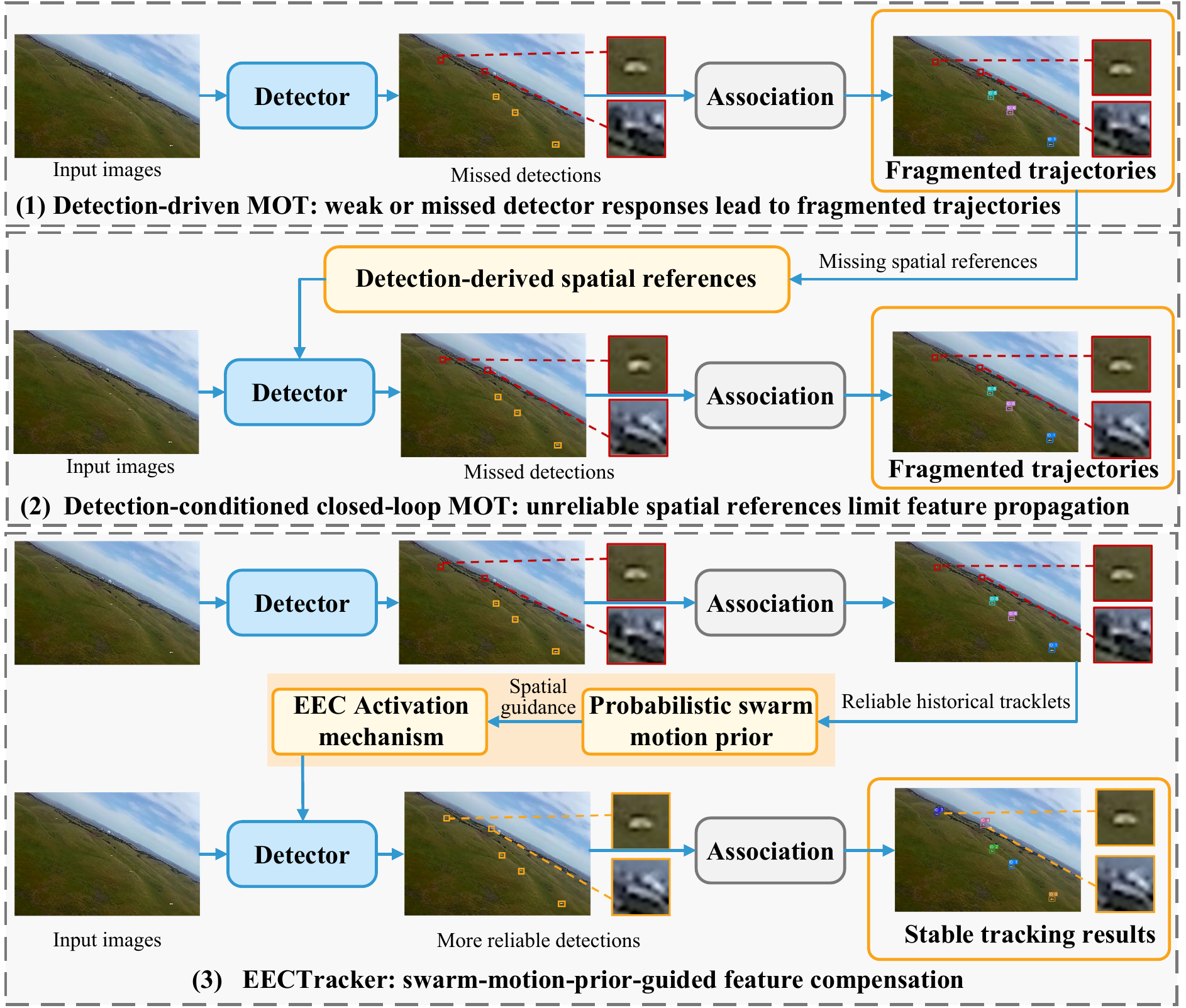}
	\caption{Conceptual comparison of tracking frameworks for airborne optical UAV swarm tracking. Detection-driven frameworks rely on detections for cross-frame association, while detection-conditioned closed-loop frameworks further exploit detection-derived spatial references for temporal feature propagation; both become vulnerable when target observations are intermittent or temporarily missing. In contrast, EECTracker introduces swarm-motion-prior-guided feature compensation, in which a probabilistic swarm motion prior provides spatial guidance and EEC Activation selectively compensates feature-consistent potential target responses for subsequent detection and association.}
	\label{fig_0}
\end{figure}

\section{Related Work}

\subsection{Airborne Optical UAV Swarm Tracking}

Recent aerial remote-sensing studies have devoted increasing attention to visual UAV tracking under challenging imaging conditions~\cite{siamSRT,siamCAP,tapTrack}. By contrast, airborne optical tracking of multiple UAVs in swarm scenarios remains relatively underexplored, with limited dedicated benchmarks and tracking algorithms. MOT-FLY~\cite{chu2023experimentalMOT-FLY} provides an early airborne multi-UAV tracking benchmark, but it mainly contains independently moving UAV targets. UAVSwarm~\cite{wang2022uavswarm} and AIRMOT~\cite{zhaochen2025visionAIRMOT+HOMA-Tracker} more directly characterize UAV swarm tracking with small long-range targets, cluttered backgrounds, homogeneous appearances, and coordinated motion patterns. Based on these benchmarks, UAVS-MOT~\cite{UAVS-MOT} improves weak UAV target representation through coordinate attention, while BELGTracker~\cite{czcBELGTracker} and HOMATracker~\cite{zhaochen2025visionAIRMOT+HOMA-Tracker} exploit geometric, motion, or association cues to improve instance discrimination. More recently, GL-DT~\cite{liu2026glGL-DT} combines spatio-temporal feature fusion and trajectory-aware tracking to improve trajectory continuity. SCT-MOT~\cite{chu2026sctmot} uses target-wise motion predictions as explicit spatial references to guide historical feature alignment and fusion. Despite these advances, intermittent or temporarily missing target observations can still interrupt trajectory continuity and identity maintenance, which remains insufficiently addressed.

\subsection{Motion-Prior-Guided Visual MOT}

Image-plane motion modeling has been widely exploited in online visual MOT to maintain temporal target information and improve cross-frame tracking. Classical recursive estimators, such as Kalman and particle filters, predict target states from historical observations~\cite{Anti-UAT}, while learning-based trajectory prediction methods further capture interaction-aware or cooperative motion patterns among multiple agents~\cite{xu2023eqmotion}. 

In visual MOT, motion cues and priors are mainly incorporated into tracking pipelines in three forms. First, displacement-based methods estimate inter-frame offsets for target localization or temporal feature propagation, as in CenterTrack~\cite{zhou2020trackingCentraltrack}, TraDeS~\cite{wu2021trackTraDeS} and IMANet~\cite{shen2023interactivelyIMANet}. Second, historical-state and query-based methods maintain temporal target information through explicit positions, track queries, or trajectory hypotheses. PPTracker~\cite{qin2025pptracker} exploits historical target positions as spatial priors, while TransTrack~\cite{sun2020transtrack} introduces track queries for cross-frame target matching. MOTR~\cite{zeng2022motr} and MOTRv3~\cite{motrv3} maintain persistent track queries across frames to couple detection and association, whereas TCT~\cite{xu2026trajectoryTCT} further introduces trajectory-level hypotheses for temporal tracking. Third, prior-fusion methods explicitly inject motion information into feature matching or attention. For example, P2FTrack~\cite{zhang2024p2ftrack} integrates detection and tracking priors with feature posteriors through prior--posterior attention.
However, most existing motion priors are constructed or updated from the observation history and state estimates of individual targets. When target observations become intermittent or temporarily missing, the available target-specific evidence is reduced, weakening the observational constraints on the corresponding motion estimates and limiting the reliability of motion-prior-guided current-frame localization and temporal feature propagation in MOT. 

\subsection{Motion-Aware Feature Modeling and Fusion}

Motion-aware feature modeling and fusion have been widely studied in video object detection and multi-object tracking. Existing approaches mainly exploit motion information in two forms. One line of work uses target-specific motion or tracking references to guide feature enhancement or detection. MP2Net~\cite{zhao2024mpMP2Net} combines mask propagation with motion prediction for satellite video object detection, while OmniTracker~\cite{wang2025omnitracker} uses tracking priors to guide current-frame detection. Another line establishes cross-frame motion correspondence to propagate and aggregate historical features. FGFA~\cite{zhu2017flowFGFA} and STSN~\cite{bertasius2018objectSTSN} perform temporal feature aggregation through motion-guided propagation or learned spatio-temporal sampling. Recent aerial and small-target methods, including DFA-MOT~\cite{zheng2025dfaDFA-MOT}, StreamFlow~\cite{10824896StreamFlow}, and LMAFormer~\cite{huang2024lmaformerLMAFormer}, further exploit optical flow, deformable alignment, or motion-aware attention for temporal feature fusion. In satellite-video MOT, FMA-Net~\cite{lu2025fmaFMA-Net} employs pixel-level flow-driven motion modeling to guide inter-frame feature fusion. Feature discrimination and reliability modeling have also been explored to identify informative responses and quantify motion uncertainty. OIFS~\cite{song2014robust} employs entropy energy to select informative appearance features for local target--background classification. ProbFlow~\cite{probflow} estimates optical-flow uncertainty from an energy-based posterior, where entropy characterizes the motion uncertainty. However, these criteria focus on appearance-feature selection or motion-field uncertainty estimation, rather than on discriminating target responses from background responses under a given motion prior.

Overall, motion-aware feature modeling exploits motion information either to provide target-specific spatial guidance for feature enhancement or to establish cross-frame correspondence for historical feature propagation and fusion. The former depends on reliable target-specific references, whereas the latter may propagate background or spatially mismatched responses together with useful target information, with the reliability of these propagated responses remaining insufficiently characterized before fusion.

\section{Method}
\subsection{Problem Formulation and Framework Overview}
\label{Problem_Formulation}

Given an online airborne optical video stream $\{I_t\}$, UAV swarm tracking aims to localize multiple UAV targets in each incoming frame and maintain their identities over time. At frame $t$, the detector produces a detection set
$\mathcal{D}_t=\{d_{j,t}\}_{j=1}^{M_t}$, where $d_{j,t}=(\mathbf{b}_{j,t},s_{j,t})$, and $\mathbf{b}_{j,t}=[x_{j,t},y_{j,t},w_{j,t},h_{j,t}]^\top$ denotes the image-plane bounding box, $s_{j,t}$ is the detection confidence. After data association, these observations are used to update the track set $\mathcal{T}_t=\{\tau_{i,t}\}_{i=1}^{N_t}$, where $\tau_{i,t}=(\mathrm{id}_i,\mathbf{b}_{i,t})$ and $\mathrm{id}_i$ denotes the object identity.

In airborne optical UAV swarm tracking, small target scales, dynamic viewpoints, and background interference can weaken current-frame target feature responses, resulting in intermittent or temporarily missing target observations and an increased risk of trajectory fragmentation and identity switches. To address this challenge, we propose EECTracker, a swarm-motion-prior-guided joint detection and tracking framework that introduces a probabilistic swarm motion prior for current-frame feature compensation. Rather than relying solely on target-specific motion estimates, the prior characterizes the shared short-term motion tendency of the swarm together with its uncertainty, thereby defining statistically reachable image-plane regions for historical feature projection. The prior is estimated online from available reliable historical tracklets and used to guide historical feature information toward these regions. EEC Activation then selectively incorporates reliable projected historical responses into the current representation before downstream detection and association.

\begin{figure*}[htbp]
	\centering
	\includegraphics[width=\textwidth]{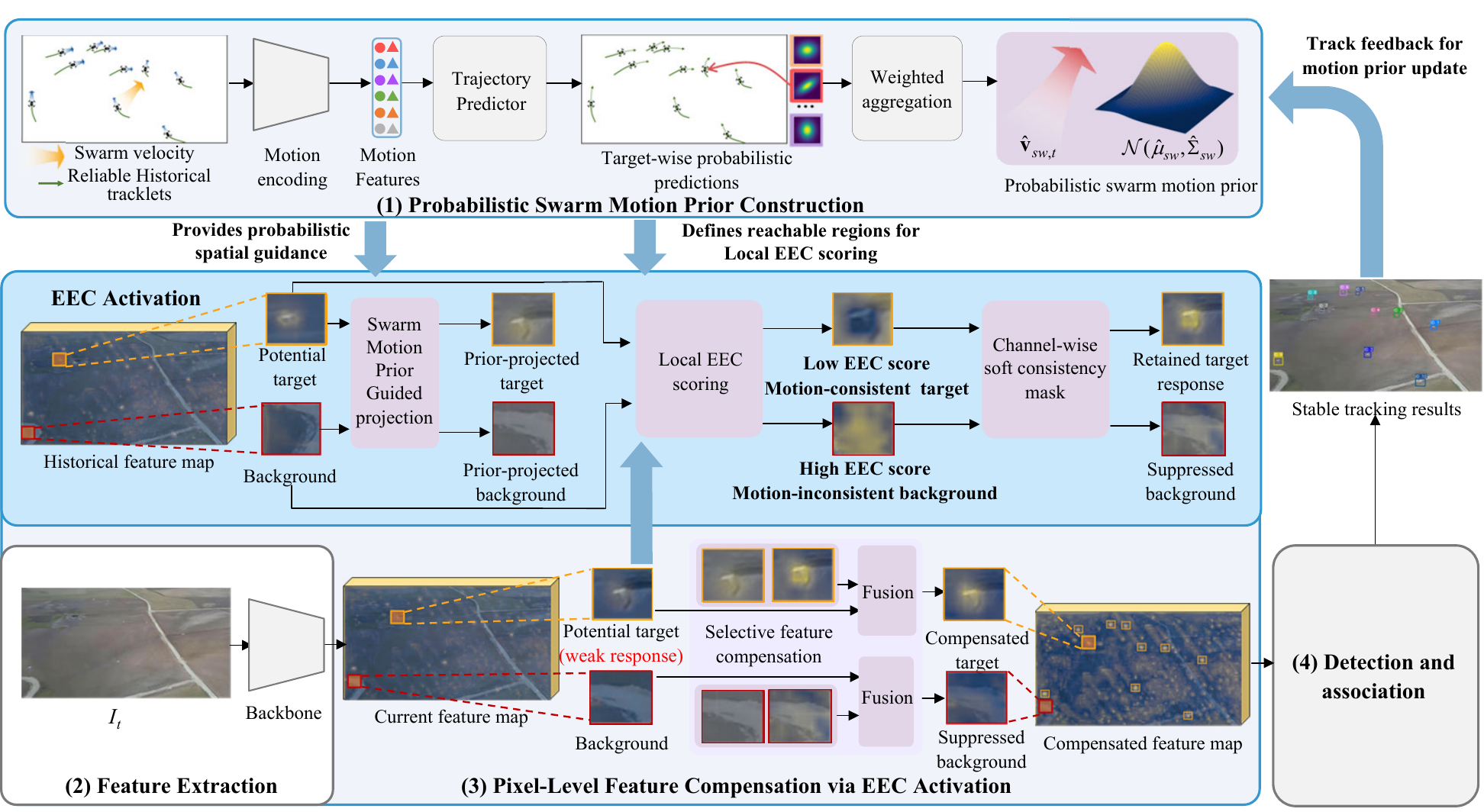}
	\caption{Overall architecture of EECTracker. The probabilistic swarm motion prior construction, pixel-level feature compensation via EEC Activation, and downstream detection and association are detailed in Sec.~\ref{SCFFC}, Sec.~\ref{Pixel-level_feature_compensation}, and Sec.~\ref{Detection_and_Tracking}, respectively.}
	\label{fig_1}
\end{figure*}

As illustrated in Fig.~\ref{fig_1}, EECTracker integrates probabilistic swarm motion prior construction, feature extraction, motion-prior-conditioned pixel-level feature compensation via EEC Activation, and downstream detection and association into a closed-loop online tracking framework.

The framework consists of four components: (1) probabilistic swarm motion prior construction, which takes the historical image-plane motion state $\mathbf{S}_{t-1}$ derived from available reliable tracklets as input and estimates a swarm-level motion prior $\mathcal{M}_t$ characterizing the statistically reachable image-plane regions of swarm targets and their associated uncertainty in the current frame; (2) feature extraction, which encodes adjacent frames $I_{t-1}$ and $I_t$ with a shared backbone to obtain multi-scale feature maps $\mathbf{F}_{t-1}$ and $\mathbf{F}_t$; (3) pixel-level feature compensation via EEC Activation, which evaluates motion-prior-conditioned consistency between historical and current feature responses, projects historical features to the current frame according to $\mathcal{M}_t$, and uses the resulting consistency to selectively weight the projected features for compensating weakened current-frame responses, yielding the compensated feature maps $\hat{\mathbf{F}}_t$; and (4) detection and tracking, which uses $\hat{\mathbf{F}}_t$ for target localization, confidence prediction, appearance embedding extraction, and cross-frame association, producing the current detection set $\mathcal{D}_t$ and updated track set $\mathcal{T}_t$. The resulting tracking states are subsequently used to update the historical motion state from $\mathbf{S}_{t-1}$ to $\mathbf{S}_t$, which is used for swarm motion prior construction at the next frame.

For adjacent frames $I_{t-1}$ and $I_t$, the shared backbone extracts multi-scale feature maps
$\mathbf{F}_{t-1}=\{\mathbf{F}^{l}_{t-1}\}_{l=1}^{m}$ and
$\mathbf{F}_{t}=\{\mathbf{F}^{l}_{t}\}_{l=1}^{m}$.
For the $l$-th feature level,
$\mathbf{F}^{l}_{t}\in\mathbf{R}^{C_l\times H_l\times W_l}$,
where $C_l$ denotes the number of channels and $(H_l,W_l)$ denotes the spatial resolution. For an input image of resolution $H_0\times W_0$, each level has a down-sampling stride $r_l$, such that
$H_l=H_0/r_l$ and $W_l=W_0/r_l$.
In our implementation, $r_l\in\{8,16,32\}$. Since UAV targets are typically small in airborne optical images, feature compensation is performed on the highest-resolution feature level $\mathbf{F}^{1}_{t}$ to preserve fine-grained target responses. Guided by $\mathcal{M}_t$, the historical feature $\mathbf{F}^{1}_{t-1}$ is projected to the current frame and selectively incorporated through EEC Activation, yielding the compensated feature map
$\hat{\mathbf{F}}^{1}_{t}\in\mathbf{R}^{C_1\times H_1\times W_1}$.
The compensated feature map replaces $\mathbf{F}^{1}_{t}$ in the multi-scale feature set to form
$\hat{\mathbf{F}}_{t}=\{\hat{\mathbf{F}}^{1}_{t},\mathbf{F}^{2}_{t},\ldots,\mathbf{F}^{m}_{t}\}$,
which is subsequently used for detection and association. The updated track set $\mathcal{T}_t$ is then fed back to update the historical motion state for swarm motion prior construction at the next frame, thereby closing the online loop from motion-prior construction to feature compensation, detection and association, and track-state update.
\subsection{Probabilistic Swarm Motion Prior Construction}
\label{SCFFC}

UAVs participating in coordinated swarm maneuvers often exhibit a shared short-term image-plane motion tendency. This collective motion property provides a transferable cue for estimating the current-frame motion of the swarm when direct observations of some targets become weak or temporarily unavailable. Based on this observation, we construct a probabilistic swarm motion prior from reliable historical tracklets to characterize the current-frame swarm motion and its uncertainty. The construction consists of five stages. First, reliable historical tracklets are selected from the online tracking results. Second, their image-plane center trajectories and velocity sequences within a temporal window are used to represent the historical motion states. Third, the historical motion states are encoded by projecting the position sequences and decomposing the velocity sequences into global swarm and residual components, yielding embedded position and motion representations for subsequent trajectory prediction. Fourth, an EqMotion-based image-plane trajectory predictor estimates the future locations of reliable tracklets and their target-wise spatial distributions. Fifth, the predicted locations are used to estimate the global swarm motion, while the target-wise spatial distributions are aggregated through motion-consistency-weighted moment matching into a compact swarm-level distribution; together, they jointly form the probabilistic swarm motion prior $\mathcal{M}_t$.

Before processing frame $t$, let $\mathcal{R}_t$ denote the subset of reliable historical tracklets used for current swarm motion modeling, and let $n_T=|\mathcal{R}_t|$. A tracklet is regarded as reliable if it has valid observations in every frame within the historical temporal window $[t-T,t-1]$ and maintains the same identity throughout the entire window. Here, $T$ denotes the temporal window length.

For each reliable tracklet $i$, its image-plane center is denoted by $\mathbf{c}_{i,t_j}=[x_{i,t_j},y_{i,t_j}]^\top$,
and its inter-frame velocity is $\mathbf{v}_{i,t_j}=\mathbf{c}_{i,t_j}-\mathbf{c}_{i,t_{j-1}}$, for $t_j\in\{t-T,\ldots,t-1\}$. The first velocity is initialized as $\mathbf{v}_{i,t-T}=\mathbf{v}_{i,t-T+1}$. Accordingly, the historical position and velocity sequences of tracklet $i$ are represented as $\mathbf{c}_i=\{\mathbf{c}_{i,t_j}\}_{t_j=t-T}^{t-1}\in\mathbb{R}^{T\times2}$ and $\mathbf{v}_i=\{\mathbf{v}_{i,t_j}\}_{t_j=t-T}^{t-1}\in\mathbb{R}^{T\times2}$, respectively. The historical image-plane motion state of the reliable tracklets within the swarm is therefore $\mathbf{S}_{t-1}=\left\{(\mathbf{c}_i,\mathbf{v}_i)
\right\}_{i=1}^{n_T}$.

To explicitly encode the shared swarm-motion information, we first estimate the average swarm velocity at each time step:
\begin{equation}
	\label{eq:historical_swarm_velocity}
	\mathbf{v}_{sw,t_j}=\frac{1}{n_T}\sum_{i=1}^{n_T}\mathbf{v}_{i,t_j},\qquad t_j\in\{t-T,\ldots,t-1\}.
\end{equation}
The corresponding historical swarm-velocity sequence is denoted as
$\mathbf{v}_{sw}=\{\mathbf{v}_{sw,t_j}\}_{t_j=t-T}^{t-1}
\in\mathbf{R}^{T\times2}$.

The position and velocity sequences are then encoded according to their respective roles in image-plane motion modeling. The position sequence describes the historical spatial evolution of each reliable tracklet, whereas the velocity sequence characterizes its short-term motion. Specifically, the position state is directly projected into the embedding space, while the velocity state is encoded by combining the global swarm velocity with the residual velocity relative to the swarm:
\begin{equation}
	\label{4-position and velocity state}
	\begin{gathered}
		\tilde{\mathbf{c}}_i = \eta_c\mathbf{c}_i\in \mathbf{R}^{T\times f}, \\
		\tilde{\mathbf{v}}_i=\eta_v(\mathbf{v}_i-\mathbf{v}_{sw})+\eta^{sw}_v\mathbf{v}_{sw}\in \mathbf{R}^{T\times f}
	\end{gathered}
\end{equation}
where $\eta_c$, $\eta_v$, $\eta^{sw}_v$ denote learnable linear projection layers, and $f$ denotes the embedding dimension.

Let $\tilde{\mathbf{C}}=\left\{\tilde{\mathbf{c}}_i\right\}_{i=1}^{n_T} \in \mathbf{R}^{n_T\times T\times f}$ and $\tilde{\mathbf{V}}=\left\{\tilde{\mathbf{v}}_i\right\}_{i=1}^{n_T} \in \mathbf{R}^{n_T\times T\times f}$ denote the embedded position and velocity states of reliable tracklets.
An EqMotion-based trajectory predictor~\cite{xu2023eqmotion} is then employed to estimate their future image-plane motion. EqMotion jointly captures the temporal motion evolution of each reliable tracklet and its interactions with other targets, enabling interaction-aware prediction for coordinated swarm motion. Since the original EqMotion produces deterministic trajectory predictions, we retain its original trajectory-prediction architecture and append a lightweight uncertainty head to provide a probabilistic spatial characterization relative to the deterministic location predicted for frame $t$: 

\begin{equation}
	\label{eq:motion_decoder}
	\left\{\hat{\mathbf{c}}_i,\left(\hat{{\mu}}_i,\hat{{\Sigma}}_i\right)\right\}_{i=1}^{n_T}=\phi_{\mathrm{traj}}\left(\tilde{\mathbf{C}},\tilde{\mathbf{V}}\right)
\end{equation}
Here, $\phi_{\mathrm{traj}}$ denotes the EqMotion-based predictor with the appended uncertainty head. For each reliable tracklet $i$, the trajectory head predicts its future image-plane locations
$\hat{\mathbf{c}}_i\in\mathbf{R}^{P\times2}$,
while the uncertainty head, implemented using a two-layer multilayer perceptron (MLP), predicts a target-wise probabilistic spatial distribution $\mathcal{N}_i(\hat\mu_i,\hat\Sigma_i)$ at frame $t$. Here, $\hat\mu_i$ denotes the mean offset relative to the predicted position $\hat{\mathbf{c}}_{i,t}$ and $\hat\Sigma_i$ denotes the corresponding spatial covariance.

Based on the prediction, we further derive the predicted velocity of each reliable swarm tracklet at frame $t$:
\begin{equation}
	\label{6-predicted velocity of each swarm tracklet}
	\begin{gathered}
		\hat{\mathbf{v}}_t=\{\hat{\mathbf{v}}_{1,t},\cdots,\hat{\mathbf{v}}_{n_T,t}\}, 
		\hat{\mathbf{v}}_{i,t}=\hat{\mathbf{c}}_{i,t}-\mathbf{c}_{i,t-1}
	\end{gathered}
\end{equation}

The predicted global swarm velocity is then estimated as
\begin{equation}
	\label{eq:predicted_swarm_velocity}
	\hat{\mathbf{v}}_{sw,t}=\frac{1}{n_T}
	\sum_{i=1}^{n_T}\hat{\mathbf{v}}_{i,t}
\end{equation}
Here, $\hat{\mathbf{v}}_{sw,t}$ summarizes the shared deterministic motion of the swarm at frame $t$.

To aggregate the target-wise probabilistic predictions according to their agreement with the shared swarm motion, we compute a directional consistency score and its normalized weight:
\begin{equation}
	\label{eq:motion_consistency_weight}
	\begin{gathered}
		s_i=\frac{\left\langle\hat{\mathbf{v}}_{i,t},\hat{\mathbf{v}}_{sw,t}\right\rangle}{\|\hat{\mathbf{v}}_{i,t}\|_2 \|\hat{\mathbf{v}}_{sw,t}\|_2+\epsilon},
		\alpha_i=\frac{\exp(\beta s_i)}{\sum_{j=1}^{n_T}\exp(\beta s_j)}
	\end{gathered}
\end{equation}
where $\epsilon$ is a small constant for numerical stability and $\beta$ controls the concentration of the normalized weights. The resulting $\alpha_i$ determines the contribution of each reliable tracklet to the swarm-level probabilistic representation.

Using these weights, we construct a compact swarm-level motion distribution by moment-matching the target-wise spatial distributions:
\begin{equation}
	\label{eq:swarm_distribution and covariances.}
	\hat{\mu}_{sw}=\sum_{i=1}^{n_T}\alpha_i \hat{\mu}_i,
	\hat{\Sigma}_{sw}=\sum_{i=1}^{n_T}\alpha_i\left[\hat{\Sigma}_i+(\hat{\mu}_i-\hat{\mu}_{sw})(\hat{\mu}_i-\hat{\mu}_{sw})^\top\right]
\end{equation}

Finally, the probabilistic swarm motion prior at time $t$ is defined as:
\begin{equation}
	\label{eq:probabilistic_swarm_motion_prior}
	\begin{split}
		\mathcal{M}_t=\{\hat{\mathbf{v}}_{sw,t},\mathcal{N}(\hat{\mu}_{sw},\hat{\Sigma}_{sw})\}
	\end{split}
\end{equation}
where $\hat{\mathbf{v}}_{sw,t}$ provides the predicted global swarm motion, and $\mathcal{N}(\hat{\mu}_{sw},\hat{\Sigma}_{sw})$ denotes a compact swarm-level probabilistic characterization of its spatial tendency and uncertainty. The $\mathcal{M}_t$ guides the subsequent EEC-based feature compensation.

\subsection{Motion-Prior-Conditioned Feature Compensation via EEC Activation}
\label{Pixel-level_feature_compensation}
Cooperative swarm UAVs often exhibit shared short-term motion patterns, leading to coherent cross-frame displacement of target regions in airborne optical imagery. In contrast, background regions and noise are generally not constrained by the same swarm-specific motion pattern, and therefore tend to exhibit weaker or less stable correspondence. This difference provides a useful cue for distinguishing potential UAV target regions and suppressing motion-inconsistent background interference.

To characterize this difference in feature space, we evaluate the cross-frame feature residuals under swarm-motion-guided correspondence. If a historical target response remains consistent with its current-frame counterpart after accounting for the shared swarm motion, the resulting cross-frame residuals are expected to exhibit both smaller magnitudes and a more spatially concentrated local distribution; in contrast, motion-inconsistent background and noise responses tend to produce larger residuals with more dispersed local distributions. We therefore use residual energy to characterize the magnitude of the cross-frame feature mismatch and residual entropy to measure the local spatial dispersion of the residual responses. Lower residual energy and entropy generally indicate stronger cross-frame motion consistency. Together, these complementary statistics provide an interpretable feature-space measure for selectively exploiting motion-consistent historical information.

\begin{figure*}[htbp]
	\centering
	\includegraphics[width=\textwidth]{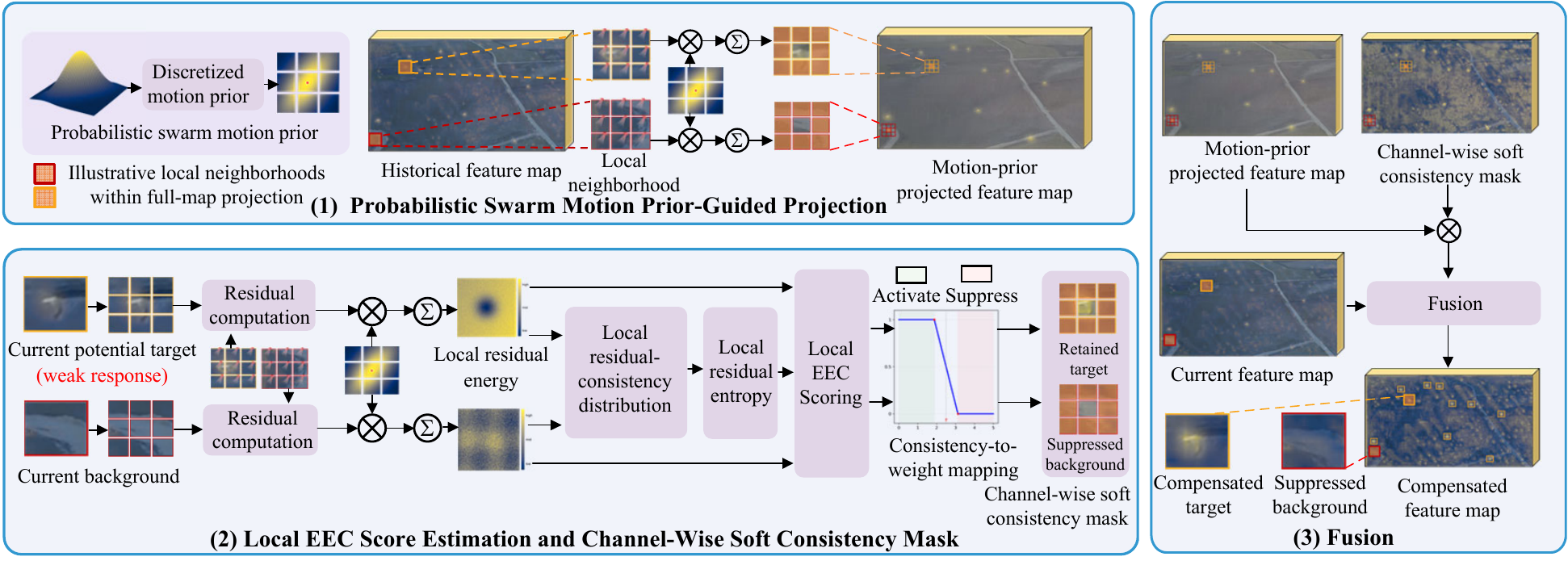}
	\caption{Pixel-level feature compensation via EEC Activation. The process consists of three stages: (1) probabilistic swarm motion prior-guided projection of historical feature responses toward the current frame; (2) Local EEC estimation from residual energy and local residual entropy, followed by channel-wise soft consistency mask construction; and (3) fusion of the EEC-weighted projected historical feature with the current feature map to obtain the compensated feature representation.}
	\label{fig_2}
\end{figure*}

Motivated by the above measurements, we propose Energy-Entropy Consistency Activation (EEC Activation), a pixel-level feature compensation mechanism guided by motion-prior-conditioned cross-frame feature consistency. As shown in Fig.~\ref{fig_2}, EEC Activation consists of three stages. First, historical features are projected toward the current frame according to the probabilistic swarm motion prior. Second, cross-frame residuals between the current response and displacement-conditioned historical responses are evaluated, from which the expected residual energy and local residual entropy are computed and combined into a Local EEC score, which is further converted into a channel-wise soft consistency mask. Third, the resulting mask is used to selectively weight the projected historical features before fusion with the current feature map, thereby compensating feature responses in potential UAV target regions and suppressing motion-inconsistent background responses.

To propagate historical features under the probabilistic swarm motion prior, we first define a cross-frame spatial mapping function $\mathcal{U}$. This function models pixel transport between adjacent frames. Specifically, given a pixel $\mathbf{p}_{t-1}$ in frame $t-1$, its current-frame correspondence is represented by a possible pixel position $\mathbf{p}_{t}$ and modeled as:

\begin{equation}
	\label{9-a cross-frame spatial mapping function}
	\begin{gathered}
		\mathbf{p}_{t}=\mathcal{U}(\mathbf{p}_{t-1}|t-1\rightarrow t)=\mathbf{p}_{t-1}+\Delta,\\
		\Delta=\hat{\mathbf{v}}_{sw,t}+\delta
	\end{gathered}
\end{equation}
where $\hat{\mathbf{v}}_{sw,t}$ represents the predicted global swarm displacement, and $\delta$ is the probabilistic spatial offset sampled from the Gaussian component $\mathcal{N}(\hat{\mu}_{sw},\hat{\Sigma}_{sw})$ of the swarm motion prior. The former provides a deterministic reference for the dominant cross-frame swarm displacement, whereas the latter characterizes the spatial uncertainty around this reference. $\Delta$ denotes the overall probabilistic displacement. In this way, the probabilistic swarm motion prior characterizes a set of statistically reachable image-plane locations for historical feature projection in the current frame, with the corresponding probability density reflecting their relative likelihood under the predicted swarm motion.

Under intermittent or temporarily missing observations, local feature alignment based on detection-derived spatial references may become unavailable or inaccurate. Instead of aligning features around detected target regions, we apply motion-prior-guided projection over the entire feature map. Let $\mathbf{F}_t=\{\mathbf{F}^1_t,\cdots,\mathbf{F}^m_t\}$ denote the multi-scale feature maps at frame $t$, where $\mathbf{F}^1_t$ is the highest-resolution level. Because UAV targets are usually extremely small in airborne optical images, feature compensation is performed only on $\mathbf{F}^1_t$ with stride $r_1=8$ to preserve fine spatial details of potential target regions while balancing effectiveness and computational efficiency. The image-plane motion parameters
are transformed into the coordinate system of this feature level:
\begin{equation}
	\hat{\mathbf{v}}^{1}_{sw,t}=\frac{\hat{\mathbf{v}}_{sw,t}}{r_1},
	\hat{{\mu}}^{1}_{sw,t}=\frac{\hat{{\mu}}_{sw,t}}{r_1},
	\hat{{\Sigma}}^{1}_{sw,t}=\frac{\hat{{\Sigma}}_{sw,t}}{r_1^{2}}.
\end{equation}

Based on the spatial mapping function $\mathcal{U}$, we project the historical feature map $\mathbf{F}^1_{t-1}$ into the current feature
coordinate system and obtain the motion-prior-projected feature map $\tilde{\mathbf{F}}^1_{t-1\rightarrow t}$. This projected feature map statistically characterizes the expected current-frame response of historical features under the swarm motion prior. 
\begin{equation}
	\label{10-motion aligned feature map}
	\begin{gathered}
		\tilde{\mathbf{F}}^1_{t-1\rightarrow t}(\mathbf{p}_{t})=\mathbb{E}_{\Delta\sim\mathcal{N}(\hat{\mathbf{v}}^1_{sw,t}+\hat{{\mu}}^1_{sw},\hat{{\Sigma}}^1_{sw})}[\mathbf{F}^1_{t-1}(\mathbf{p}_{t}-\Delta)]=\\
		\int_{\delta}\mathbf{F}^1_{t-1}(\mathbf{p}_{t}-\delta-\hat{\mathbf{v}}^1_{sw,t})\mathcal{N}(\delta;\hat{\mu}^1_{sw},\hat{\Sigma}^1_{sw})d\delta
	\end{gathered}
\end{equation}

For a target-associated location whose image-plane motion follows the swarm motion prior, the historical feature response sampled at its inverse-mapped position is expected to be close to the corresponding current-frame response. This approximate motion-prior-conditioned feature consistency can be expressed as: 
\begin{equation}
	\label{11-approximate motion-consistency across adjacent frames}
	\begin{gathered}
		\mathbf{F}^1_{t-1}(\mathcal{U}_{\Delta}^{-1}(\mathbf{p}_{t}))\approx \mathbf{F}^1_{t}(\mathbf{p}_{t})
	\end{gathered}
\end{equation}

Accordingly, we further define a channel-wise expected feature residual over the displacement distribution. For pixel $\mathbf{p}_t$ and channel $c$, the channel residual under a displacement $\Delta$ is defined as: 
\begin{equation}
	\label{12- channel residual under a displacement hypothesis }
	\begin{gathered}
		\mathbf{r}_c(\mathbf{p}_t;\Delta)={\mathbf{F}}^1_{t-1, c}(\mathcal{U}_{\Delta}^{-1}(\mathbf{p}_t))-\mathbf{F}^1_{t,c}(\mathbf{p}_t)=\\
		\mathbf{F}^1_{t-1,c}(\mathbf{p}_{t}-\Delta)-\mathbf{F}^1_{t,c}(\mathbf{p}_{t})
	\end{gathered}
\end{equation}
where $\mathbf{F}_{t,c}^{1}$ and $\mathbf{F}_{t-1,c}^{1}$ denote the feature responses of the current and historical feature maps at channel $c$, respectively. The residual $\mathbf{r}_c(\mathbf{p}_t;\Delta)$ measures the cross-frame feature mismatch between the current response and the historical response sampled at the corresponding inverse-mapped position. Based on the channel residual, we then define the channel-wise residual energy as the expected squared feature residual over the displacement distribution induced by the swarm motion prior:
\begin{equation}
	\label{13-channel-wise residual energy}
	\begin{gathered}
		E_c(\mathbf{p}_{t})=\mathbb{E}_{\Delta\sim\mathcal{N}(\hat{\mathbf{v}}^1_{sw,t}+\hat{{\mu}}^1_{sw},\hat{{\Sigma}}^1_{sw})}[\mathbf{r}_c(\mathbf{p}_t;\Delta)^2]
	\end{gathered}
\end{equation}

This residual energy measures the expected magnitude of cross-frame feature mismatch under plausible swarm motion displacements. Target regions consistent with the swarm motion typically exhibit lower residual energy, whereas motion-inconsistent background regions generally produce larger residual energy.

On the discrete pixel grid, we approximate the above channel-wise residual energy by discretizing the displacement distribution within a local neighborhood $\Omega(\mathbf{p}_t)$. Since feature maps are defined on discrete pixel locations, the continuous expectation in Eq. \ref{13-channel-wise residual energy} is implemented by a normalized weighted summation over the discrete sampling locations. Specifically, for a pixel $\mathbf{p}_t$, we consider candidate sampling locations $\mathbf{u}\in\Omega(\mathbf{p}_t)$ and use the corresponding feature
responses to construct a discrete approximation of the
displacement-induced feature distribution.
\begin{equation}
	\label{14-discrete approximation of the displacement-induced feature distribution}
	\begin{gathered}
		w(\mathbf{u}|\mathbf{p}_t)=\frac{\mathcal{N}\big(\mathbf{p}_t-\mathbf{u}; \hat{\mu}^1_{sw},\hat{\Sigma}^1_{sw}\big)}
		{\sum_{\mathbf{u}'\in\Omega(\mathbf{p}_t)} \mathcal{N}\big(\mathbf{p}_t-\mathbf{u}'; \hat{\mu}^1_{sw},\hat{\Sigma}^1_{sw}\big)}
	\end{gathered}
\end{equation}

Accordingly, the discrete approximation of the motion-prior-projected historical feature at $\mathbf{p}_t$ is given by:
\begin{equation}
	\label{15-aligned feature}
	\begin{gathered}
		\tilde{\mathbf{F}}^1_{t-1\rightarrow t}(\mathbf{p}_t)=\sum_{\mathbf{u}\in\Omega(\mathbf{p}_t)}w(\mathbf{u}|\mathbf{p}_t) \mathbf{F}^1_{t-1}(\mathbf{u}-\hat{\mathbf{v}}^1_{sw,t})
	\end{gathered}
\end{equation}

The corresponding channel-wise expected residual energy is approximated as:
\begin{equation}
	\label{16-channel-wise expected residual energy}
	\begin{gathered}
		E_c(\mathbf{p}_{t})=\sum_{\mathbf{u}\in\Omega(\mathbf{p}_{t})}w(\mathbf{u}|\mathbf{p_t})(\mathbf{F}^1_{t-1,c}(\mathbf{u}-\hat{\mathbf{v}}^1_{sw,t})-\mathbf{F}^1_{t,c}(\mathbf{p}_t))^2
	\end{gathered}
\end{equation}

Residual energy characterizes the magnitude of the cross-frame feature mismatch but does not describe the spatial concentration of local consistency responses. We further normalize the channel-wise residual energy within the local neighborhood to obtain a local residual-consistency distribution:
\begin{equation}
	\label{17-local probability distribution of residual energy}
	\begin{gathered}
		\pi_c(\mathbf{q}_t|\mathbf{p}_t)=\frac{\exp(-E_c(\mathbf{q}_t))}{\sum_{\mathbf{k}_t\in \Omega(\mathbf{p}_t)}\exp(-E_c(\mathbf{k}_t))},\mathbf{q}_t\in \Omega(\mathbf{p}_t)
	\end{gathered}
\end{equation}

The channel-wise residual entropy is then defined as:
\begin{equation}
	\label{18-channel-wise residual entropy}
	\begin{gathered}
		S_c(\mathbf{p}_t)=-\sum_{\mathbf{q}_t\in\Omega(\mathbf{p_t})}\pi_c(\mathbf{q}_t|\mathbf{p}_t)\log\pi_c(\mathbf{q}_t|\mathbf{p}_t)
	\end{gathered}
\end{equation}

Here, residual entropy measures the local spatial dispersion of residual-consistency responses. For target regions consistent with the swarm motion prior, low-residual responses are usually more spatially concentrated, leading to lower entropy. In contrast, motion-inconsistent background regions tend to produce more dispersed residual responses and therefore higher entropy.

Based on the channel-wise residual energy and residual entropy, we construct a Local EEC score to jointly characterize the magnitude and local spatial dispersion of the cross-frame feature residuals. Specifically, for pixel $\mathbf{p}_t$ and channel $c$, the Local EEC score is defined as:
\begin{equation}
	\label{19-Local EEC score }
	\begin{gathered}
		G_c(\mathbf{p}_t)=E_c(\mathbf{p}_t)+\lambda_s S_c(\mathbf{p}_t)
	\end{gathered}
\end{equation}
where $\lambda_s$ balances the contributions of the residual energy and entropy terms. A lower $G_c(\mathbf{p}_t)$ indicates stronger motion-prior-conditioned consistency between the historical and current feature responses in channel $c$, whereas a higher value indicates weaker consistency.

We then construct a learnable channel-wise soft consistency mask $M(\mathbf{p}_t)\in \mathbf{R}^C_1$ to adaptively activate potential target features based on $G_c$. For channel $c$, the mask is defined as:
\begin{equation}
	\label{20-the mask}
	\begin{gathered}
		M_c(\mathbf{p}_t)=\sigma(\alpha_c(\tau_c-G_c(\mathbf{p}_t)))
	\end{gathered}
\end{equation}
where $\tau_c\in \mathbf{R}$ and $\alpha_c > 0$ denote the learnable consistency threshold and scale factor for channel $c$, respectively. Both parameters can be optimized end-to-end during training. $\sigma(\cdot)$ denotes the HardSigmoid activation function. With this design, each channel can adaptively modulate its activation according to its response to motion-prior-conditioned consistency, enabling finer-grained control of historical feature contributions.

We then apply the mask $M(\mathbf{p}_t)$ to selectively weight the motion-prior-projected historical feature and fuse it with the current-frame feature, producing the swarm-motion-guided compensated feature map $\hat{\mathbf{F}}^1_t$:
\begin{equation}
	\label{21-swarm motion guided recovered feature map}
	\begin{gathered}
		\hat{\mathbf{F}}^1_t=\phi_{fu}([M\odot \tilde{\mathbf{F}}^1_{t-1\rightarrow t};\mathbf{F}^1_t]) + \mathbf{F}^1_t
	\end{gathered}
\end{equation}
where $\phi_{fu}$ denotes a $3\times 3$ convolutional fusion layer, $[\cdot,\cdot]$ denotes channel-wise concatenation, and $\odot$ denotes the Hadamard product. This operation retains historical responses with stronger motion-prior-conditioned consistency while attenuating inconsistent responses, thereby enabling pixel-level compensation of potential target features under intermittent or temporarily missing observations.

Finally, the compensated feature map $\hat{\mathbf{F}}^1_t$ replaces $\mathbf{F}^1_t$ in the multi-scale feature set and is fed into the subsequent detection and tracking branches for target detection and association.

\subsection{Detection and Tracking Integration}
\label{Detection_and_Tracking}
To integrate probabilistic swarm motion prior construction and feature compensation into an online tracking loop, we adopt an existing swarm tracking framework, HOMATracker\cite{zhaochen2025visionAIRMOT+HOMA-Tracker} as the downstream detection and tracking framework. The compensated feature set $ \hat{\mathbf F}_t $ is fed into this framework for detection, appearance embedding extraction, and cross-frame association, producing the current tracking result $\mathcal{T}_t$ for subsequent motion state update.

Specifically, $\hat{\mathbf{F}}_t$ is processed by two parallel branches for target detection and appearance embedding extraction. The detection branch predicts the detection set $\mathcal{D}_t$, including bounding boxes and confidence scores. In parallel, the appearance branch extracts target-level embeddings $\mathbf{E}_t$ from $ \hat{\mathbf{F}}_t$ for identity discrimination and data association.

During association, we follow the multi-frame homogeneous association strategy of HOMATracker and construct the association cost using appearance similarity and motion consistency. The current detections $\mathcal{D}_t$ are then associated with historical tracklets $\mathcal{T}_{t-1}$ to obtain the updated track set $\mathcal{T}_t$. This process is performed within a sliding temporal window to improve association stability.

The resulting track set $\mathcal{T}_t$ is then used to update the historical motion state $\mathbf{S}_t$. Before processing frame $t+1$, tracklets that remain continuously observed and maintain consistent identities throughout the temporal window are selected to form the reliable set $\mathcal{R}_{t+1}$, which is then used to construct the next-frame probabilistic swarm motion prior $\mathcal{M}_{t+1}$. This feedback process closes the online loop among motion-prior construction, feature compensation, detection, and tracking, enabling the tracker to continuously exploit reliable historical swarm motion information and thereby improve tracking robustness under intermittent or temporarily missing observations.

\section{Experiments}
\subsection{Datasets}
We evaluate EECTracker on two representative public benchmarks for airborne optical UAV swarm tracking: AIRMOT and UAVSwarm. AIRMOT is a simulation-based benchmark, whereas UAVSwarm is collected from real flight scenarios.

AIRMOT is an open-source simulation benchmark for airborne optical swarm UAV tracking. It contains 8 highly dynamic RGB video sequences with a resolution of $1920\times 1080$ and a total of 7,844 frames. A major challenge of AIRMOT is the extremely small apparent target scale: more than 97\% of targets are smaller than $32\times 32$ pixels, which leads to weak visual responses and intermittent detections. Each sequence typically contains 5--16 UAV instances undergoing coordinated maneuvers and formation changes.

UAVSwarm is a real-world UAV swarm tracking benchmark collected from flight scenarios. It contains 72 video sequences with a total of 12,598 annotated frames, covering 13 different scenes and more than 19 UAV types, and is officially split into 36 training sequences and 36 testing sequences. Compared with AIRMOT, UAVSwarm presents more complex background interference, including urban buildings, clouds, and ground clutter, together with multiple viewpoints such as top-down and horizontal views. It also includes representative swarm behaviors such as formation reconfiguration and cooperative maneuvers.

We follow the official train/test splits for all experiments, with models trained on the training set and evaluated on the test set. 
To train the probabilistic swarm motion prior construction module, we further construct auxiliary trajectory-prediction sequences from the original training annotations of both benchmarks. For each video sequence, we extract continuous trajectory segments of length $L=20$ using a spatio-temporal sliding window. Following a standard trajectory-prediction setting, the first $L_{obs}=8$ frames are used as historical observations and the remaining $L_{pred}=12$ frames are used for future-motion supervision. During sequence construction, the frame-level temporal alignment among UAV tracklets is preserved, so that the predictor can jointly learn individual image-plane motion evolution and short-term swarm coordination patterns. This procedure yields two auxiliary trajectory-prediction subsets, denoted as AIRMOT-STP (Swarm Trajectory Prediction) and UAVSwarm-STP, respectively. These subsets are used only for pretraining the swarm motion-prior construction module and do not alter the official tracking benchmark splits.  

\subsection{Evaluation Metrics}
We evaluate EECTracker from three aspects: detection accuracy, tracking performance, and runtime efficiency. For detection evaluation, we report Average Precision at an IoU threshold of 0.5 (AP50). For tracking evaluation, we report Multiple Object Tracking Accuracy (MOTA), Identification F1 Score (IDF1), Higher Order Tracking Accuracy (HOTA), and the number of identity switches (IDSW). Runtime efficiency is measured by the end-to-end inference speed in frames per second (FPS)~\cite{MOTReview}. 

AP50 measures the detection accuracy by integrating the precision--recall curve at an IoU threshold of 0.5:
\begin{equation}
	\label{eq:AP50}
	\mathrm{AP}_{50}=\int_{0}^{1} P(R) dR
\end{equation}
where $P$ and $R$ denote the precision and recall evaluated at an IoU threshold of 0.5, respectively.

MOTA measures the overall tracking accuracy by accounting for false positives, false negatives, and identity switches:
\begin{equation}
	\label{eq:MOTA}
	\mathrm{MOTA}=1-\frac{\sum_t\left(\mathrm{FN}_t+\mathrm{FP}_t+\mathrm{IDSW}_t\right)}{\sum_t \mathrm{GT}_t}
\end{equation}
where $\mathrm{FP}_t$, $\mathrm{FN}_t$, $\mathrm{IDSW}_t$, and $\mathrm{GT}_t$ denote the numbers of false positives, false negatives, identity switches, and ground-truth objects at frame $t$, respectively.

IDF1 evaluates identity preservation over the entire sequence and is defined as
\begin{equation}
	\label{eq:IDF1}
	\mathrm{IDF1}=\frac{2\cdot\mathrm{IDTP}}{2\cdot\mathrm{IDTP}+\mathrm{IDFP}+\mathrm{IDFN}},
\end{equation}
where $\mathrm{IDTP}$, $\mathrm{IDFP}$, and $\mathrm{IDFN}$ denote the numbers of true-positive, false-positive, and false-negative identity assignments, respectively. IDSW directly counts the total number of identity switches:
\begin{equation}
	\label{eq:IDSW}
	\mathrm{IDSW}=\sum_t \mathrm{IDSW}_t.
\end{equation}

HOTA jointly evaluates detection and association quality by averaging the corresponding scores over a set of localization thresholds, and is computed as
\begin{equation}
	\label{eq:HOTA}
	\mathrm{HOTA}=\frac{1}{|\mathcal{A}|}\sum_{\alpha\in\mathcal{A}}
	\sqrt{\mathrm{DetA}_{\alpha}\mathrm{AssA}_{\alpha}},
\end{equation}
where $\mathrm{DetA}_{\alpha}$ and $\mathrm{AssA}_{\alpha}$ denote the detection and association accuracy at threshold $\alpha$, respectively, and $\mathcal{A}$ denotes the set of localization thresholds used for evaluation.

Finally, runtime efficiency is evaluated using the end-to-end inference speed in frames per second (FPS). Higher AP50, MOTA, IDF1, HOTA, and FPS indicate better performance, whereas lower IDSW is preferred.
\subsection{Implementation Details}
We train and evaluate all models on a single NVIDIA RTX 3090 GPU.

We first pretrain the probabilistic swarm motion prior construction (PSMP) module. An EqMotion-based \cite{xu2023eqmotion} trajectory predictor is employed to predict target-wise future image-plane locations, together with a lightweight uncertainty head for estimating the corresponding local spatial uncertainty. The predictor takes $L_{\mathrm{obs}}=8$ historical frames as input and outputs two complementary motion representations: a deterministic future trajectory over the subsequent $P=L_{\mathrm{pred}}=12$ frames and a target-wise Gaussian spatial offset distribution parameterized by $(\hat{{\mu}}_i,\hat{{\Sigma}}_i)$. The former provides the predicted image-plane motion reference, while the latter characterizes the corresponding local spatial uncertainty. We optimize this module using stochastic gradient descent (SGD) with an initial learning rate of $5\times10^{-4}$, momentum of 0.9, and weight decay of $1\times10^{-4}$ for 60 epochs.

To supervise trajectory prediction, we use a mean squared position error loss:
\begin{equation}
	\label{26-mean squared error.}
	\begin{gathered}
		\mathcal{L}_{pos}=\frac{1}{n_TP}\sum^{P}_{t_j=1}\sum^{n_T}_{i=1}||\hat{\mathbf{c}}_{i}[t_j]-\mathbf{c}^{*}_{i}[t_j]||^2_2
	\end{gathered}
\end{equation}
where $\mathbf{c}^{*}_{i}$ denotes the ground truth future trajectory of target $i$ over the next $P$ frames, and $\hat{\mathbf{c}}_{i}$ denotes the corresponding prediction.

To supervise the  probabilistic spatial prediction at frame $t$, we construct a local spatial reference by averaging the ground-truth positions within a temporal neighborhood:
\begin{equation}
	\label{eq:local_average_position}
	\bar{\mathbf{c}}^{*}_{i,t}=\frac{1}{2P_u+1}\sum_{q=-P_u}^{P_u}\mathbf{c}^{*}_{i,t+q},
\end{equation}
where $P_u=3$ determines the temporal support of the local spatial reference. The uncertainty head predicts the mean $\hat{{\mu}}_i$ and covariance $\hat{{\Sigma}}_i$ of the target-wise spatial offset and is trained using the negative log-likelihood (NLL) loss:
\begin{equation}
	\label{eq:uncertainty_loss}
	\mathcal{L}_{\mathrm{nll}}=-\frac{1}{n_T}\sum_{i=1}^{n_T}\log
	\mathcal{N}\left(\bar{\mathbf{c}}^{*}_{i,t}-\hat{\mathbf{c}}_{i,t};\hat{{\mu}}_i,\hat{{\Sigma}}_i\right)
\end{equation}

The overall training loss of the PSMP module is:
\begin{equation}
	\label{28-The overall training loss of the swarm-consistent flow field module}
	\begin{gathered}
		\mathcal{L}_{traj}=\lambda_1\mathcal{L}_{pos}+\lambda_2\mathcal{L}_{nll}
	\end{gathered}
\end{equation}
where $\lambda_1=1$ and $\lambda_2=0.01$ are fixed for all experiments.

After pretraining the PSMP module, we train the full MOT framework. We adopt HOMATracker as the base framework and integrate both the PSMP module and EEC Activation into it. During training, all input images are resized while preserving the original aspect ratio and padded to an input size of $1088\times1088$, and the network is optimized using SGD with an initial learning rate of 0.002, momentum of 0.9, and weight decay of $1\times10^{-4}$.

A sliding temporal window of $T=8$ frames is used to maintain the historical motion states for probabilistic swarm motion prior construction. During full-framework training, the prior is constructed from historical tracklets and used to guide EEC-based historical feature compensation. The pretrained PSMP module remains frozen, while EEC Activation is optimized end-to-end together with the downstream tracking framework through the original detection and association losses. Specifically, we follow the multi-branch loss design of HOMATracker, including bounding-box regression, category prediction, objectness prediction, and appearance association losses, collectively denoted as $\mathcal{L}_{\mathrm{track}}$. No additional supervision is introduced for EEC Activation.

For EEC Activation, the probabilistic displacement distribution is discretized over a fixed rectangular neighborhood around each query pixel. The local neighborhood size is set to $3\times3$ on AIRMOT and $7\times7$ on UAVSwarm based on the ablation results. The balance factor of the Local EEC score is fixed to $\lambda_s=0.1$. The learnable channel-wise gating parameters $\alpha_c$ and $\tau_c$ are initialized as 1.0 and 0.8, respectively, and constrained to $[0.001, 2.0]$ during training for numerical stability. These settings are fixed for all sequences within each dataset.

During inference, the framework processes the video stream online and predicts swarm motion prior using a sliding window of length $T=8$ and stride 1. In the first $T$ frames, the framework performs conventional detection and tracking to initialize target trajectories and historical motion-state buffer. From frame $T+1$ onward, tracklets satisfying the reliable-tracklet selection rule are used by the PSMP module to construct the probabilistic swarm motion prior $\mathcal{M}_t$. If no reliable tracklet is available at frame $t$, PSMP and EEC Activation are bypassed, and the original current-frame features $\mathbf{F}_t$ are retained. Otherwise, EEC Activation projects historical features toward the current frame under $\mathcal{M}_t$, computes the Local EEC score, and applies the resulting channel-wise soft consistency mask to weight the projected historical features before fusion with the current-frame feature map. The compensated feature set $\hat{\mathbf{F}}_t$ is subsequently fed into the downstream detection and tracking branches.

All input images are resized with their original aspect ratios preserved and then padded to $1088 \times 1088$. The confidence threshold for candidate detection is 0.01, and non-maximum suppression (NMS) is applied with a threshold of 0.7. The detection branch outputs the candidate target set $\mathcal{D}_t$ for the current frame. For runtime evaluation, we report the end-to-end inference speed of the complete online framework in FPS, including image preprocessing, feature extraction, probabilistic swarm motion prior construction, EEC-based feature compensation, downstream detection, association and post-processing.  

For data association, we follow the multi-frame homogeneous association strategy of HOMATracker and maintain the appearance features and motion states of each tracklet within the sliding window. Detections in each frame are first divided into high-confidence and low-confidence sets according to their confidence scores. High-confidence detections are assigned to existing tracklets using the Hungarian algorithm with a cost matrix computed from appearance similarity and motion difference. Unmatched high-confidence detections are initialized as new tracklets. Low-confidence detections are associated only with unmatched tracklets from the previous frame based on spatial IoU to reduce false associations.

\subsection{Comparison with State-of-the-Art Methods}
To comprehensively evaluate the effectiveness of EECTracker for airborne optical UAV swarm tracking, we compare it with representative methods from three mainstream MOT paradigms: tracking-by-detection (TBD), joint detection and tracking (JDT), and transformer-based end-to-end tracking. The compared methods include DeepSORT\cite{deepsort}, ByteTrack\cite{bytetrack}, OC-SORT\cite{ocsort}, Hybrid-SORT\cite{hybridsort}, BELGTracker\cite{czcBELGTracker}, HOMATracker\cite{zhaochen2025visionAIRMOT+HOMA-Tracker}, FairMOT\cite{fairmot}, UAVS-MOT\cite{UAVS-MOT}, MOTRv3\cite{motrv3} and SCT-MOT\cite{chu2026sctmot}.

To ensure a fair comparison, we control irrelevant factors other than the tracking framework as much as possible. For TBD methods, we employ YOLOX-X as the unified front-end detector, and all methods are trained and evaluated under the same train/test split and input resolution. For JDT and end-to-end methods, we preserve their original framework designs and follow their standard training and inference pipelines, while using the same data splits, input resolution, and evaluation protocol whenever applicable. 

\begin{table*}[!t]
	\centering
	\caption{Quantitative comparison with state-of-the-art methods on AIRMOT and UAVSwarm. $\uparrow$/$\downarrow$: higher/lower is better. Best results on each dataset are shown in bold.}
	\label{tab:table1}
	\begin{tabular*}{\textwidth}{@{\hspace{2pt}}l@{\extracolsep{\fill}}lccccc@{\hspace{2pt}}}
		\toprule
		\textbf{Methods} & \textbf{Publication} & \textbf{MOTA(\%)$\uparrow$} & \textbf{IDF1(\%)$\uparrow$} & \textbf{HOTA(\%)$\uparrow$} & \textbf{IDSW$\downarrow$} & \textbf{FPS}$\uparrow$ \\
		\toprule
		AIRMOT             &       &       &       &       &       &       \\ \midrule
		FairMOT\cite{fairmot}            & IJCV 2021 & 18.19 & 17.41 & 17.56 & 476   & 35.2  \\
		DeepSORT\cite{deepsort}           & ICIP 2017 & 19.60 & 20.45 & 19.42 & 2841  & 33.6  \\
		OC-SORT\cite{ocsort}             & CVPR 2023 & 20.88 & 23.27 & 22.07 & 522   & \textbf{40.9} \\
		ByteTrack\cite{bytetrack}          & ECCV 2022 & 24.35 & 25.72 & 23.56 & 1707  & 39.2  \\
		MOTRv3\cite{motrv3}             & ArXiv 2023 & 23.52 & 24.89 & 22.59 & 1805  & 20.6  \\
		Hybrid-SORT\cite{hybridsort}         & AAAI 2024 & 29.01 & 24.08 & 21.70 & 652   & 28.9  \\
		BELGTracker\cite{czcBELGTracker}        & Acta Aero. Sin. (CN) 2024 & 29.27 & 23.55 & 23.47 & 1144  & 34.6  \\
		HOMATracker\cite{zhaochen2025visionAIRMOT+HOMA-Tracker}        & CJA 2025 & 30.08 & 29.31 & 25.55 & 506   & 20.0  \\
		SCT-MOT\cite{chu2026sctmot}            & TAES 2026 & 32.30 & 30.15 & 25.54 & 476   & 21.8  \\
		EECTracker         & ours & \textbf{36.19} & \textbf{31.94} & \textbf{26.15} & \textbf{400} & 22.4 \\ \midrule
		UAVSwarm           &       &       &       &       &       &       \\ \midrule
		FairMOT\cite{fairmot}            & IJCV 2021 & 67.7  & 73.2  & 59.2  & 590   & 35.2  \\
		DeepSORT\cite{deepsort}           & ICIP 2017 & 61.2  & 70.3  & 58.8  & 221   & 33.6  \\
		OC-SORT\cite{ocsort}             & CVPR 2023 & 75.7  & 81.8  & 64.7  & 709   & \textbf{40.9} \\
		ByteTrack\cite{bytetrack}          & ECCV 2022 & 65.0  & 76.5  & 60.6  & 67    & 39.2  \\
		MOTRv3\cite{motrv3}             & ArXiv 2023 & 62.3  & 71.9  & 57.9  & 72    & 20.6  \\
		Hybrid-SORT\cite{hybridsort}         & AAAI 2024 & 77.4  & 80.1  & 62.8  & 459   & 28.9  \\
		UAVS-MOT\cite{UAVS-MOT}        & Acta Aero. Sin. (CN) 2024 & 73.4  & 76.1  & 65.8  & 740   & 34.6  \\
		HOMATracker\cite{zhaochen2025visionAIRMOT+HOMA-Tracker}        & CJA 2025 & 79.2  & 87.1  & 67.0  & 58    & 20.3  \\
		SCT-MOT\cite{chu2026sctmot}            & TAES 2026 & 81.90 & 88.45 & 68.56 & 56 & 22.3  \\
		EECTracker         & ours & \textbf{84.71} & \textbf{90.19} & \textbf{69.16} & \textbf{54} & 22.9 \\
		\bottomrule
	\end{tabular*}
\end{table*}

The quantitative results on AIRMOT and UAVSwarm are shown in Table~\ref{tab:table1}. EECTracker achieves the best MOTA, IDF1, HOTA, and IDSW on both benchmarks, demonstrating consistent improvements in overall tracking accuracy and identity preservation. On AIRMOT, compared with SCT-MOT, EECTracker improves MOTA, IDF1, and HOTA by 3.89, 1.79, and 0.61 percentage points, respectively, while reducing IDSW by 76. On UAVSwarm, the corresponding improvements are 2.81, 1.74, and 0.60 percentage points, respectively, with IDSW further reduced from 56 to 54. 

Overall, EECTracker yields stable improvements across both datasets, with more pronounced benefits on AIRMOT. This trend is consistent with the more severe small-target observation degradation in AIRMOT, where current-frame UAV responses are more frequently weakened or intermittently missed. In terms of efficiency, EECTracker maintains online inference speed and remains comparable to the most relevant joint tracking baselines such as HOMATracker and SCT-MOT, indicating that the framework provides a favorable balance between tracking performance and online computational efficiency.

\begin{figure*}[htbp]
	\centering
	\includegraphics[width=\textwidth]{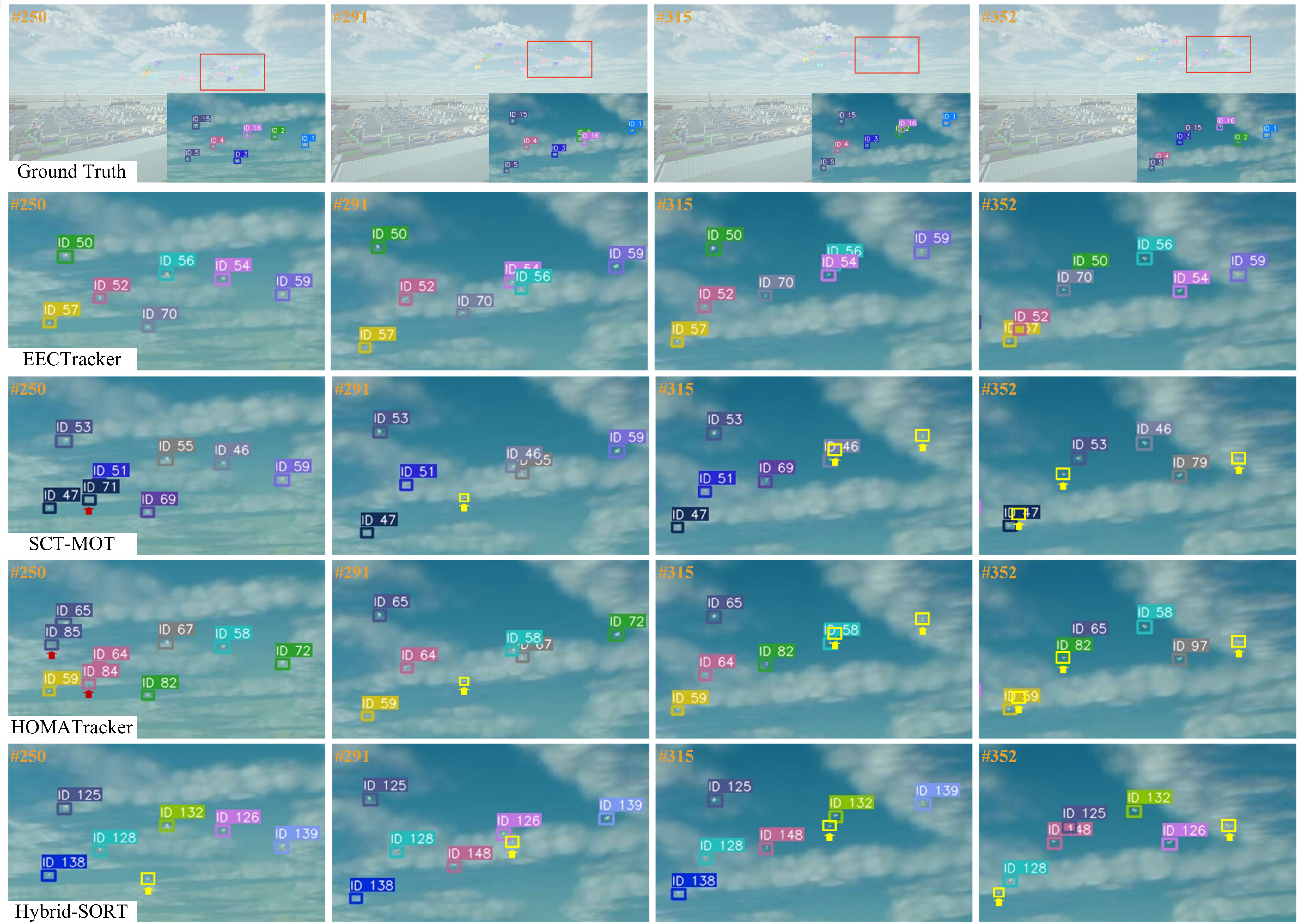}
	\caption{Qualitative comparison on AIRMOT under small target scales, weak visual responses, and cloud-background interference. Different tracked objects are marked by bounding boxes with different colors. Red arrows denote false positives, while yellow boxes and arrows indicate missed detections.}
	\label{fig_4}
\end{figure*}

\begin{figure*}[htbp]
	\centering
	\includegraphics[width=\textwidth]{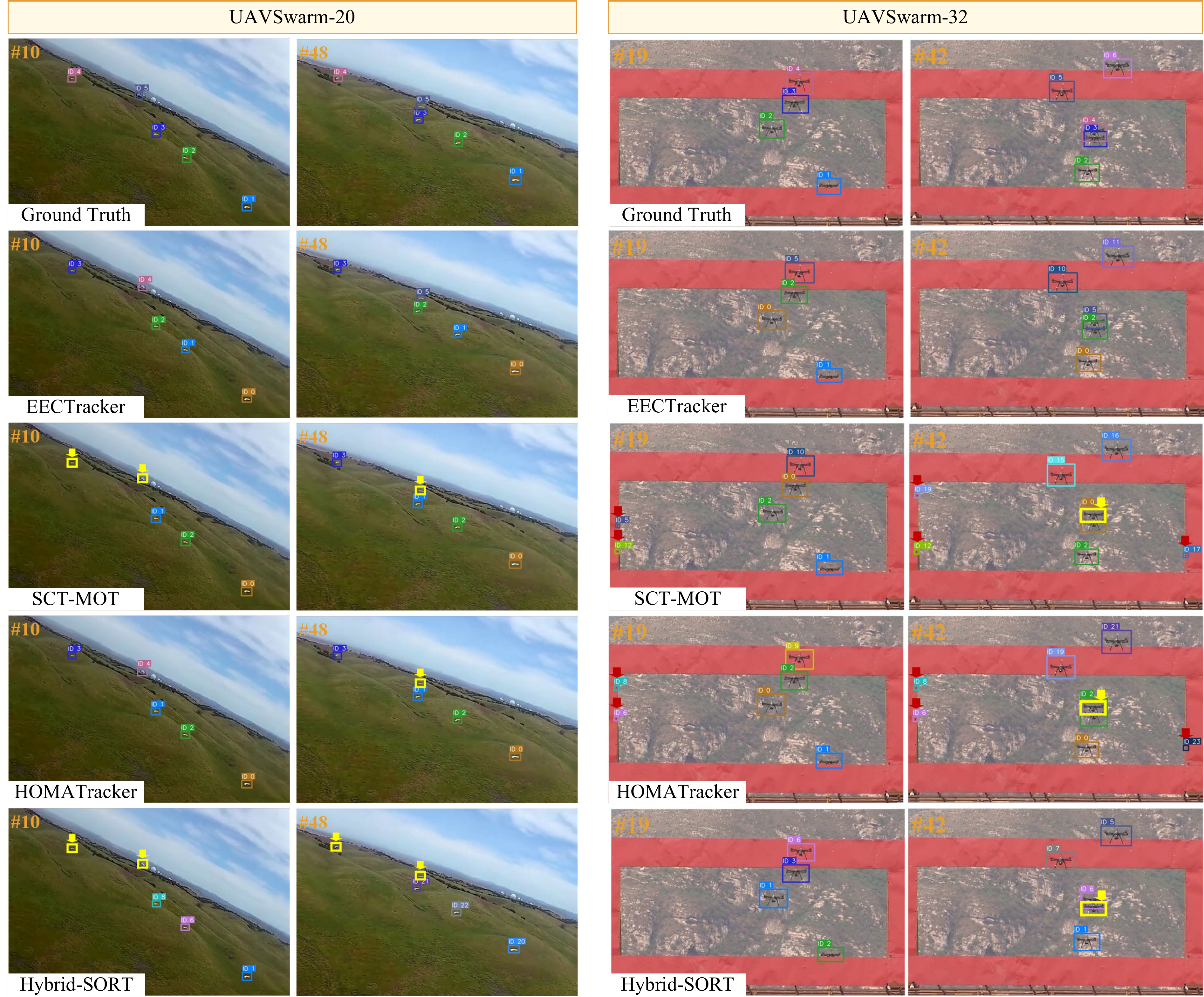}
	\caption{Qualitative comparison on UAVSwarm under weak target responses, complex background clutter, and partial occlusion. Different tracked objects are marked by bounding boxes with different colors. Red arrows denote false positives, while yellow boxes and arrows indicate missed detections.}
	\label{fig_5}
\end{figure*}

To complement the quantitative evaluation, we present qualitative comparisons on the AIRMOT and UAVSwarm datasets in Figs.~\ref{fig_4} and \ref{fig_5}. For visual clarity, we compare EECTracker with the most relevant competitors, including SCT-MOT as the strongest prior method, HOMATracker as the direct base framework, and Hybrid-SORT as a representative strong TBD baseline when applicable. In the figures, different tracked targets are marked by bounding boxes with different colors; red arrows denote false positives, while yellow boxes and arrows indicate missed detections. The selected examples are taken from representative sequences involving weak target responses, cluttered backgrounds, and severe appearance degradation to highlight the robustness differences among the compared methods.

In the AIRMOT-02 sequence, targets are difficult to perceive between frames 250 and 352 due to small scale, weak appearance, and cloud-background interference. Under these conditions, the compared methods exhibit varying false alarms and missed detections. At frame 250, for example, SCT-MOT and HOMATracker produce one to two false positives. At frame 352, multiple targets are missed by SCT-MOT, HOMATracker, and Hybrid-SORT. In contrast, EECTracker maintains visibly more stable detection and tracking results over the selected interval, with fewer false alarms, fewer missed detections, and more consistent identities. 
On UAVSwarm, we further analyze two representative clips. In UAVSwarm-20, the combination of small target size and complex ground clutter causes noticeable missed detections for competing methods; for example, HOMATracker misses one target at frame 48. In UAVSwarm-32, at frame 42, SCT-MOT and HOMATracker produce false alarms for nearly half of the targets because of interference from background border structures. Meanwhile, partial occlusion among adjacent UAVs causes missed detections for multiple competing methods. By comparison, EECTracker maintains more stable detection and tracking results for all targets under these challenging conditions.
These qualitative observations are consistent with the quantitative results and further illustrate the role of the probabilistic swarm motion prior and EEC Activation. The motion prior provides statistically reachable image-plane regions for historical feature projection, while EEC Activation performs feature compensation according to motion-prior-conditioned feature consistency, strengthening useful historical responses in potential UAV target regions while attenuating motion-inconsistent responses. Together, these mechanisms improve detection reliability and tracking continuity under weak and intermittent target observations in airborne optical imagery.

\subsection{Effectiveness of Probabilistic Swarm Motion Prior-Guided Feature Compensation}

To evaluate the effectiveness of the proposed swarm-motion-prior-guided feature compensation strategy, we conduct an ablation study under a unified HOMATracker-based framework. All variants share the same backbone, detection branch, appearance branch, and association strategy. For variants requiring motion prediction, the same PSMP module is used, so that the comparison focuses on the feature compensation strategy. We compare four variants: (1) Baseline, which uses HOMATracker without feature compensation; (2) Temporal feature concatenation, which directly concatenates and fuses adjacent-frame features; (3) Explicit detector-location-guided compensation, which uses detector-provided target locations to guide feature compensation and is implemented using the representative IMA fusion mechanism from IMANet~\cite{shen2023interactivelyIMANet}; and (4) Swarm Motion-prior-guided feature compensation (EECTracker). The results are reported in Table~\ref{tab:feature_compensation_ablation}.

\begin{figure}[!t]
	\centering
	\subfloat[AIRMOT]{
		\includegraphics[width=0.98\linewidth]{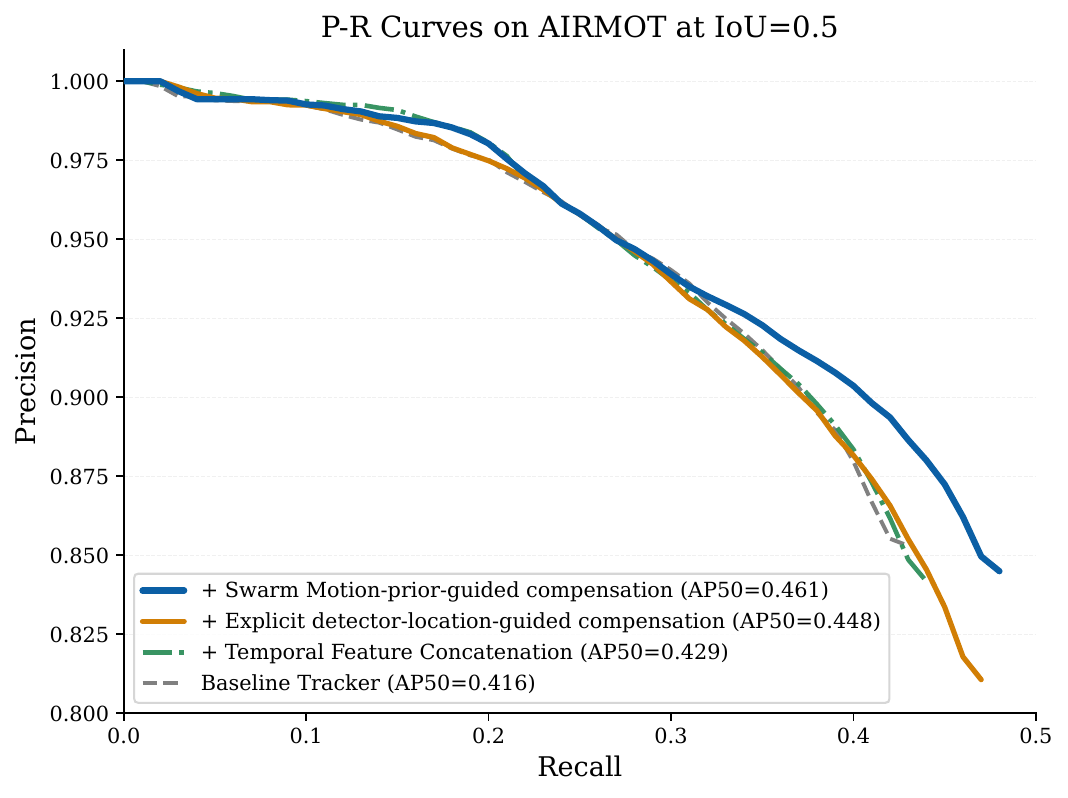}
		\label{AIRMOT_PR_curve}
	}
	
	\vspace{-2mm}
	
	\subfloat[UAVSwarm]{
		\includegraphics[width=0.98\linewidth]{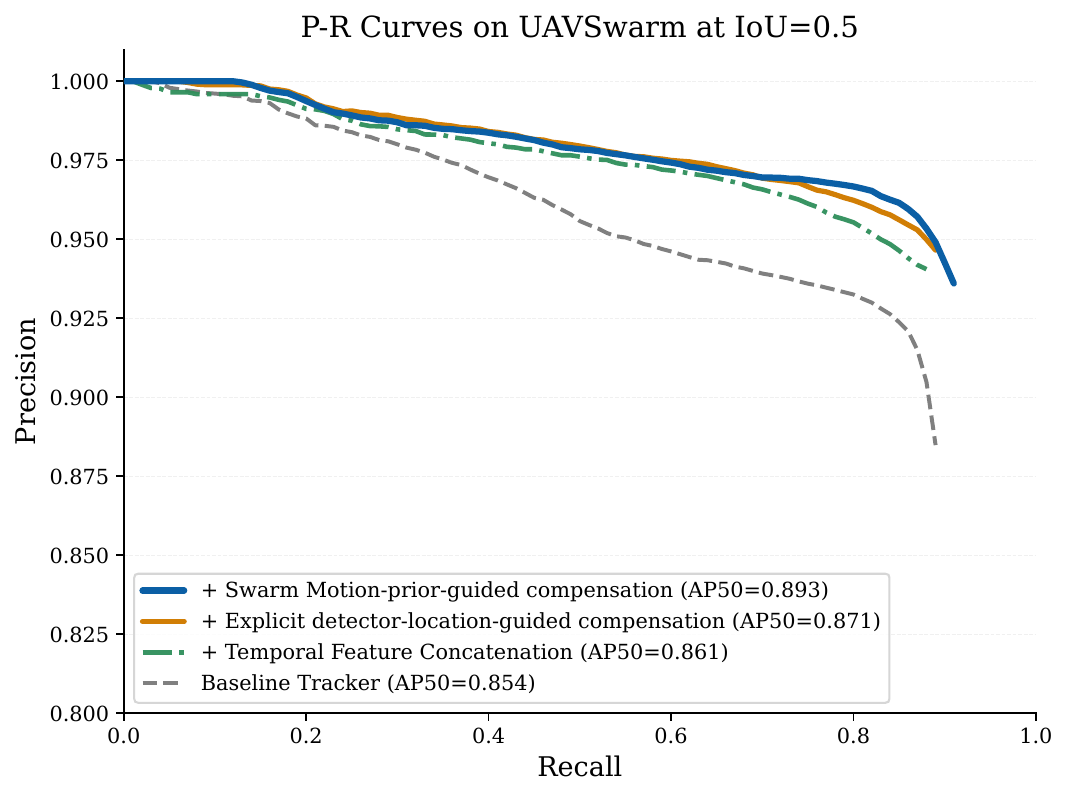}
		\label{UAVSwarm_PR_curve}
	}
	\vspace{-1mm}
	\caption{Precision-recall curves of different feature compensation strategies on AIRMOT and UAVSwarm at IoU=0.5.}
	\label{fig_3}
	\vspace{-2mm}
\end{figure}

As shown in Table~\ref{tab:feature_compensation_ablation}, EECTracker achieves the best AP50, MOTA, IDF1, and HOTA on both AIRMOT and UAVSwarm. On AIRMOT, compared with explicit detector-location-guided compensation, EECTracker improves AP50, MOTA, IDF1, and HOTA by 1.30, 4.35, 3.56, and 1.49 percentage points, respectively, and reduces IDSW by 89. On UAVSwarm, EECTracker improves AP50, MOTA, IDF1, and HOTA over explicit detector-location-guided compensation by 2.20, 4.66, 2.56, and 1.87 percentage points, while achieving the same lowest IDSW of 54.

The comparison among the compensation strategies further reveals their different behaviors. Temporal feature concatenation improves AP50 and MOTA over the baseline, but it brings limited or even negative gains in IDF1 and HOTA, especially on AIRMOT. This indicates that simply introducing adjacent-frame features does not necessarily provide reliable temporal information for downstream tracking. Explicit detector-location-guided compensation improves AP50, IDF1, HOTA, and IDSW over temporal feature concatenation, but slightly decreases MOTA on both datasets, suggesting that detector-provided target locations can make feature alignment more discriminative when these locations are reliable, while their dependence on current target locations may constrain compensation under weak or intermittent observations. In contrast, EECTracker combines probabilistic swarm-level spatial guidance with motion-prior-conditioned feature consistency, resulting in more consistent improvements across detection and tracking metrics.

SCT-MOT~\cite{chu2026sctmot} is a closely related motion-prior-guided UAV swarm tracking method built on SMTP for target-wise trajectory prediction and TG-STFF for trajectory-guided historical feature fusion. To compare this target-specific deterministic guidance with the swarm-motion-prior-guided feature compensation in EECTracker, we keep the remaining HOMATracker-based framework unchanged and evaluate three variants: (1) SCT-MOT with SMTP and TG-STFF; (2) PSMP+TG-STFF, which replaces SMTP with PSMP while retaining trajectory-guided fusion; and (3) PSMP+EEC Activation, which further replaces TG-STFF with consistency-guided feature compensation. The results are reported in Table~\ref{tab:sct_component}.

As shown in Table~\ref{tab:sct_component}, when coupled with the same TG-STFF fusion module, PSMP improves AP50, MOTA, and IDF1 on both datasets and reduces IDSW relative to SMTP, while maintaining comparable HOTA. This indicates that PSMP provides more effective current-frame motion guidance than the target-wise deterministic predictions used in SCT-MOT. More importantly, under the same PSMP, EEC Activation consistently outperforms TG-STFF across all reported metrics. On AIRMOT, it further improves MOTA, IDF1, and HOTA by 1.27, 1.09, and 0.58 percentage points, respectively, while reducing IDSW by 62; on UAVSwarm, the corresponding gains are 2.26, 0.98, and 0.88 percentage points, while reducing IDSW by 1. These results show that motion-prior-conditioned feature consistency enables more effective historical feature utilization for downstream detection and tracking than trajectory-guided fusion alone.

The precision-recall curves in Fig.~\ref{fig_3} provide complementary detection-level evidence. On both datasets, the compared methods perform similarly in the low-recall region, where detections are mainly determined by relatively salient targets. As recall increases, the detector includes more challenging UAV instances, such as weak or ambiguous targets. In this high-recall region, EECTracker preserves higher precision than the competing variants, indicating that swarm-motion-prior-guided feature compensation is more effective in maintaining reliable detections for challenging targets. This advantage emerges earlier on AIRMOT, becoming evident beyond approximately 0.3 recall, which is consistent with its smaller target scales and more frequent weak target responses. On UAVSwarm, the advantage mainly emerges beyond approximately 0.7 recall, suggesting that the compared variants remain close on easy targets, while EECTracker becomes more effective as increasingly challenging targets are included. These results are consistent with the AP50 improvements in Table~\ref{tab:feature_compensation_ablation} and further support the effectiveness of the proposed compensation strategy under weak and intermittent target observations.

\begin{table*}[!t]
	\centering
	\setlength{\tabcolsep}{2.0pt}
	\renewcommand{\arraystretch}{1.0}
	\caption{Ablation of feature compensation strategies on AIRMOT and UAVSwarm. ``Det. loc." denotes whether feature compensation requires detector-provided target locations. AP50, MOTA, IDF1, and HOTA are reported as percentages.}
	\label{tab:feature_compensation_ablation}
	\begin{tabular*}{\textwidth}{@{\extracolsep{\fill}}p{4.7cm}c*{5}{c}*{5}{c}@{}}
		\toprule
		\multirow{2}{*}{\textbf{Variant}}
		& \multirow{2}{*}{\textbf{Det. loc.}}
		& \multicolumn{5}{c}{\textbf{AIRMOT}}
		& \multicolumn{5}{c}{\textbf{UAVSwarm}} \\
		\cmidrule(lr){3-7} \cmidrule(lr){8-12}
		&
		& \textbf{AP50$\uparrow$}
		& \textbf{MOTA$\uparrow$}
		& \textbf{IDF1$\uparrow$}
		& \textbf{HOTA$\uparrow$}
		& \textbf{IDSW$\downarrow$}
		& \textbf{AP50$\uparrow$}
		& \textbf{MOTA$\uparrow$}
		& \textbf{IDF1$\uparrow$}
		& \textbf{HOTA$\uparrow$}
		& \textbf{IDSW$\downarrow$} \\
		\midrule
		Baseline Tracker
		& No
		& 41.60 & 30.08 & 29.31 & 25.55 & 506
		& 85.40 & 79.20 & 87.10 & 67.00 & 58 \\
		
		+ Temporal Feature Concatenation
		& No
		& 42.90 & 32.23 & 28.06 & 24.01 & 497
		& 86.10 & 80.61 & 87.35 & 67.15 & 64 \\
		
		\begin{tabular}[c]{@{}l@{}}+ Explicit Detector-Location-\\ Guided Compensation\end{tabular}
		& Yes
		& 44.80 & 31.84 & 28.38 & 24.66 & 489
		& 87.10 & 80.05 & 87.63 & 67.29 & \textbf{54} \\
		
		\begin{tabular}[c]{@{}l@{}}+ Swarm Motion Prior-\\ Guided Compensation\end{tabular}
		& No
		& \textbf{46.10} & \textbf{36.19} & \textbf{31.94} & \textbf{26.15} & \textbf{400}
		& \textbf{89.30} & \textbf{84.71} & \textbf{90.19} & \textbf{69.16} & \textbf{54} \\
		\bottomrule
	\end{tabular*}
\end{table*}

\begin{table*}[!t]
	\centering
	\setlength{\tabcolsep}{1.8pt}
	\renewcommand{\arraystretch}{1.0}
	\caption{Component-level comparison with SCT-MOT on AIRMOT and UAVSwarm under the same HOMATracker-based framework.}
	\label{tab:sct_component}
	\begin{tabular*}{\textwidth}{@{\extracolsep{\fill}}p{2.6cm}cc*{5}{c}*{5}{c}@{}}
		\toprule
		\multirow{2}{*}{\textbf{Variant}}
		& \multirow{2}{*}{\textbf{Motion Prior}}
		& \multirow{2}{*}{\textbf{Feature Fusion}}
		& \multicolumn{5}{c}{\textbf{AIRMOT}}
		& \multicolumn{5}{c}{\textbf{UAVSwarm}} \\
		\cmidrule(lr){4-8} \cmidrule(lr){9-13}
		&
		&
		& \textbf{AP50$\uparrow$}
		& \textbf{MOTA$\uparrow$}
		& \textbf{IDF1$\uparrow$}
		& \textbf{HOTA$\uparrow$}
		& \textbf{IDSW$\downarrow$}
		& \textbf{AP50$\uparrow$}
		& \textbf{MOTA$\uparrow$}
		& \textbf{IDF1$\uparrow$}
		& \textbf{HOTA$\uparrow$}
		& \textbf{IDSW$\downarrow$} \\
		\midrule
		
		SCT-MOT
		& SMTP
		& TG-STFF
		& 45.30 & 32.30 & 30.15 & 25.54 & 476
		& 86.30 & 81.90 & 88.45 & 68.56 & 56 \\
		
		PSMP + TG-STFF
		& PSMP
		& TG-STFF
		& 45.40 & 34.92 & 30.85 & 25.57 & 462
		& 87.40 & 82.45 & 89.21 & 68.28 & 55 \\
		
		EECTracker
		& PSMP
		& EEC Activation
		& \textbf{46.10} & \textbf{36.19} & \textbf{31.94} & \textbf{26.15} & \textbf{400}
		& \textbf{89.30} & \textbf{84.71} & \textbf{90.19} & \textbf{69.16} & \textbf{54} \\
		
		\bottomrule
	\end{tabular*}
\end{table*}

\subsection{Cross-Tracker Plug-in Evaluation of the Proposed Feature Compensation}
\label{sec:cross_tracker_plugin}

To evaluate whether the proposed feature compensation strategy can be integrated into different tracking frameworks, we conduct plug-in experiments on ByteTrack, OC-SORT, and FairMOT. ByteTrack and OC-SORT are representative TBD methods, while FairMOT is a representative JDT framework. For ByteTrack and OC-SORT, PSMP and EEC Activation are inserted into the detector used by the tracker, between the backbone/neck features and the detection head. The compensated feature representation is then fed into the original detection head, while the subsequent association pipeline remains unchanged. For FairMOT, the two components are inserted between the shared feature map and the detection/ReID heads, so that feature compensation is performed before joint detection and appearance prediction. In all cases, reliable historical tracklets produced by the original tracker are used to construct the probabilistic swarm motion prior, while the original detection heads, ReID branches, matching logic, and association strategies are kept unchanged. All experiments follow the same data splits, input resolution, training protocol, and inference settings as the corresponding base trackers.

As shown in Table~\ref{tab:cross_tracker_plugin}, adding PSMP and EEC Activation improves all three trackers on both datasets, indicating that the proposed feature compensation strategy is not restricted to the HOMATracker-based implementation of EECTracker. For example, the ByteTrack enhanced with the proposed feature compensation improves MOTA, IDF1, and HOTA by 2.90, 0.93, and 0.61 percentage points on AIRMOT, and by 3.18, 2.26, and 0.82 percentage points on UAVSwarm, respectively. For FairMOT, the corresponding improvements are 2.53, 4.57, and 3.33 percentage points on AIRMOT, and 1.61, 1.32, and 0.83 percentage points on UAVSwarm. Consistent gains are also observed on OC-SORT, while IDSW decreases for all three trackers on both datasets. These results suggest that, when reliable historical tracklets are available, the proposed feature compensation strategy can be integrated into different MOT frameworks and provide consistent performance gains without modifying their original association strategies.

\begin{table*}[!t]
\centering
\caption{Cross-tracker plug-in evaluation of EEC-based feature compensation on AIRMOT and UAVSwarm.``Comp." indicates whether PSMP and EEC Activation are enabled.}
\label{tab:cross_tracker_plugin}
\begin{tabular*}{\textwidth}{@{\extracolsep{\fill}}llcccccccc@{}}
\toprule
\multirow{2}{*}{\textbf{Tracker}} 
& \multirow{2}{*}{\textbf{Comp.}}
& \multicolumn{4}{c}{\textbf{AIRMOT}}
& \multicolumn{4}{c}{\textbf{UAVSwarm}} \\
\cmidrule(lr){3-6} \cmidrule(lr){7-10}
& 
& \textbf{MOTA$\uparrow$} 
& \textbf{IDF1$\uparrow$} 
& \textbf{HOTA$\uparrow$} 
& \textbf{IDSW$\downarrow$}
& \textbf{MOTA$\uparrow$} 
& \textbf{IDF1$\uparrow$} 
& \textbf{HOTA$\uparrow$} 
& \textbf{IDSW$\downarrow$} \\
\midrule
\multirow{2}{*}{ByteTrack} 
& $\times$ 
& 24.35 & 25.72 & 23.56 & 1707 
& 65.00 & 76.50 & 60.60 & 67 \\
& $\checkmark$ 
& \textbf{27.25} & \textbf{26.65} & \textbf{24.17} & \textbf{1254} 
& \textbf{68.18} & \textbf{78.76} & \textbf{61.42} & \textbf{65} \\
\midrule
\multirow{2}{*}{OC-SORT} 
& $\times$ 
& 20.88 & 23.27 & 22.07 & 522 
& 75.70 & 81.80 & 64.70 & 709 \\
& $\checkmark$ 
& \textbf{21.53} & \textbf{23.31} & \textbf{22.15} & \textbf{481} 
& \textbf{76.44} & \textbf{82.33} & \textbf{64.88} & \textbf{578} \\
\midrule
\multirow{2}{*}{FairMOT} 
& $\times$ 
& 18.19 & 17.41 & 17.56 & 476 
& 67.70 & 73.20 & 59.20 & 590 \\
& $\checkmark$ 
& \textbf{20.72} & \textbf{21.98} & \textbf{20.89} & \textbf{425} 
& \textbf{69.31} & \textbf{74.52} & \textbf{60.03} & \textbf{561} \\
\bottomrule
\end{tabular*}
\end{table*}
\subsection{Effectiveness and Robustness of Probabilistic Swarm Motion Prior Construction}
\label{sec:psmp_analysis}

The PSMP module constructs the image-plane probabilistic swarm motion prior used by EEC Activation for motion-prior-guided feature compensation. We evaluate PSMP from three aspects: (1) the effectiveness of probabilistic swarm motion prior construction; (2) robustness to historical tracklet inputs, including tracklet availability and the train--inference gap; and (3) the effect of prior representation.

\subsubsection{Effectiveness of Swarm-Aware Probabilistic Prior Construction}

\begin{table*}[!t]
	\centering
	\setlength{\tabcolsep}{1.8pt}
	\renewcommand{\arraystretch}{1.0}
	\caption{Ablation of probabilistic motion prior construction on AIRMOT and UAVSwarm. ``S.A.", ``Gen.", and ``KF" denote swarm-aware modeling, prior generator, and Kalman filter, respectively.}
	\label{tab:prior_construction_ablation}
	\begin{tabular*}{\textwidth}{@{\extracolsep{\fill}}p{3.95cm}cc*{5}{c}*{5}{c}@{}}
		\toprule
		\multirow{2}{*}{\textbf{Variant}}
		& \multirow{2}{*}{\textbf{S.A.}}
		& \multirow{2}{*}{\textbf{Gen.}}
		& \multicolumn{5}{c}{\textbf{AIRMOT}}
		& \multicolumn{5}{c}{\textbf{UAVSwarm}} \\
		\cmidrule(lr){4-8} \cmidrule(lr){9-13}
		&
		&
		& \textbf{AP50$\uparrow$}
		& \textbf{MOTA$\uparrow$}
		& \textbf{IDF1$\uparrow$}
		& \textbf{HOTA$\uparrow$}
		& \textbf{IDSW$\downarrow$}
		& \textbf{AP50$\uparrow$}
		& \textbf{MOTA$\uparrow$}
		& \textbf{IDF1$\uparrow$}
		& \textbf{HOTA$\uparrow$}
		& \textbf{IDSW$\downarrow$} \\
		\midrule
		Kalman Probabilistic Prior
		& $\times$ & KF
		& 44.30 & 34.86 & 30.49 & 25.56 & 468
		& 87.50 & 83.95 & 89.06 & 68.43 & 73 \\
		
		Individual Learned Prior
		& $\times$ & Learned
		& 45.20 & 35.68 & 30.86 & 25.47 & 447
		& 88.20 & 84.23 & 89.57 & 68.68 & 67 \\
		
		Swarm-Aware Learned Prior
		& \checkmark & Learned
		& \textbf{46.10} & \textbf{36.19} & \textbf{31.94} & \textbf{26.15} & \textbf{400}
		& \textbf{89.30} & \textbf{84.71} & \textbf{90.19} & \textbf{69.16} & \textbf{54} \\
		\bottomrule
	\end{tabular*}
\end{table*}

The probabilistic swarm motion prior constructed by PSMP serves as the common motion reference for both historical feature projection and Local EEC score computation. Its quality directly affects the reliability of subsequent feature compensation. Therefore, we evaluate the effectiveness of the proposed prior construction strategy by comparing three probabilistic motion-prior variants while keeping the downstream feature compensation module and the detection-and-tracking framework unchanged: (1) Kalman probabilistic prior, which  estimates target-wise displacement and covariance using a classical Kalman filter. (2) Individual learned probabilistic prior, which removes the swarm-aware motion encoding from PSMP and learns each target’s displacement distribution and uncertainty from its individual historical motion. (3) Swarm-aware learned probabilistic prior (EECTracker),  which uses PSMP to model swarm-aware motion relations and target-wise uncertainty for constructing the swarm motion prior. For all variants, the resulting target-wise probabilistic estimates are aggregated using the same weighting strategy to construct the final motion prior for downstream feature compensation. The results are reported in Table~\ref{tab:prior_construction_ablation}.

As shown in Table~\ref{tab:prior_construction_ablation}, the proposed swarm-aware learned probabilistic prior achieves the best overall performance on both AIRMOT and UAVSwarm. Compared with the individual learned prior, it improves AP50, MOTA, IDF1, and HOTA by 0.90, 0.51, 1.08, and 0.68 percentage points on AIRMOT, respectively, while reducing IDSW by 47. On UAVSwarm, the corresponding improvements are 1.10, 0.48, 0.62, and 0.48 percentage points, respectively, with 13 fewer ID switches. These results indicate that, under the same probabilistic prediction and aggregation framework, incorporating shared short-term swarm motion information provides more effective probabilistic guidance for current-frame feature compensation than modeling individual target motion independently.

Meanwhile, the individual learned prior also outperforms the Kalman probabilistic prior on most metrics. On AIRMOT, it improves AP50, MOTA, and IDF1 by 0.90, 0.82, and 0.37 percentage points, respectively, and reduces IDSW by 21, while HOTA remains comparable. On UAVSwarm, it improves AP50, MOTA, IDF1, and HOTA by 0.70, 0.28, 0.51, and 0.25 percentage points, respectively, with 6 fewer ID switches. These results suggest that learned target-wise motion and uncertainty provide more effective probabilistic guidance for feature compensation than classical target-wise recursive filtering.

\subsubsection{Robustness to Historical Tracklet Inputs}

Since PSMP constructs the swarm motion prior from reliable historical tracklets, we analyze its robustness from two aspects: reliable-tracklet availability and the train--inference gap of historical tracklet inputs. The former analyzes how PSMP behaves when the number of reliable tracklets decreases, while the latter examines the gap between online predicted tracklets and annotated historical tracklets.

\textbf{Tracklet availability: }
To evaluate the sensitivity of PSMP to reduced reliable-tracklet availability, we vary the proportion of reliable tracklets used for motion-prior construction. During inference, the overall EECTracker framework and inference pipeline are kept unchanged. The 100\% setting uses all tracklets satisfying the reliable-tracklet selection rule within the 8-frame historical window, whereas the 75\%, 50\%, and 25\% settings randomly retain the corresponding proportions of these tracklets. The 0\% setting disables PSMP and reduces to the baseline tracker without the swarm motion prior.

\begin{table}[!t]
	\centering
	\setlength{\tabcolsep}{2.0pt}
	\renewcommand{\arraystretch}{1.05}
	\caption{Sensitivity analysis of PSMP to reliable tracklet availability.}
	\label{tab:tracklet_availability}
	\begin{tabular*}{\columnwidth}{@{\extracolsep{\fill}}lcccccc@{}}
		\toprule
		\textbf{Dataset} 
		& \textbf{Ratio}
		& \textbf{AP50$\uparrow$}
		& \textbf{MOTA$\uparrow$}
		& \textbf{IDF1$\uparrow$}
		& \textbf{HOTA$\uparrow$}
		& \textbf{IDSW$\downarrow$} \\
		\midrule
		\multirow{5}{*}{AIRMOT}
		& 100\% & \textbf{46.10} & \textbf{36.19} & \textbf{31.94} & \textbf{26.15} & \textbf{400} \\
		& 75\%  & 45.60 & 35.86 & 31.33 & 26.06 & 405 \\
		& 50\%  & 44.80 & 35.27 & 30.92 & 25.97 & 413 \\
		& 25\%  & 43.30 & 34.63 & 29.75 & 25.24 & 458 \\
		& 0\%   & 41.60 & 30.08 & 29.31 & 25.55 & 506 \\
		\midrule
		\multirow{5}{*}{UAVSwarm}
		& 100\% & \textbf{89.30} & \textbf{84.71} & \textbf{90.19} & \textbf{69.16} & \textbf{54} \\
		& 75\%  & 89.10 & 84.29 & 89.62 & 68.90 & 62 \\
		& 50\%  & 88.50 & 83.35 & 89.23 & 68.54 & 69 \\
		& 25\%  & 86.60 & 82.82 & 89.08 & 67.15 & 73 \\
		& 0\%   & 85.40 & 79.20 & 87.10 & 67.00 & 58 \\
		\bottomrule
	\end{tabular*}
\end{table}

As shown in Table~\ref{tab:tracklet_availability}, tracking performance degrades gradually as the proportion of reliable tracklets decreases from 100\% to 25\%. On AIRMOT, the 75\% and 50\% settings remain close to the full setting, with MOTA drops of only 0.33 and 0.92 percentage points, respectively. Even with only 25\% reliable tracklets, PSMP still improves AP50, MOTA, and IDF1 over the 0\% baseline. A similar trend is observed on UAVSwarm, where the 50\% setting retains clear performance gains, and the 25\% setting still outperforms the baseline in AP50, MOTA, IDF1, and HOTA. These results indicate that PSMP remains effective with limited reliable-tracklet availability, although its benefit gradually decreases as fewer historical tracklets are available for motion-prior construction.

\textbf{Train--inference gap of historical tracklet inputs: }
PSMP is trained with annotated trajectories but constructs the motion prior from online tracklets during inference. To evaluate this train--inference gap, we compare two settings: (1) Online PSMP, the default EECTracker setting, which constructs the motion prior from online tracklets generated by the tracking pipeline; and (2) GT-tracklet PSMP, which constructs the motion prior from the corresponding ground-truth trajectories within the same historical window. The GT setting is used only for sensitivity analysis and does not use future-frame annotations. All other settings are kept unchanged.

\begin{table}[!t]
	\centering
	\setlength{\tabcolsep}{1.8pt}
	\renewcommand{\arraystretch}{1.05}
	\caption{Train--inference gap analysis of historical tracklet inputs for PSMP.}
	\label{tab:gt_online_tracklet_gap}
	\begin{tabular*}{\columnwidth}{@{\extracolsep{\fill}}llccccc@{}}
		\toprule
		\textbf{Dataset} 
		& \textbf{Input}
		& \textbf{AP50$\uparrow$}
		& \textbf{MOTA$\uparrow$}
		& \textbf{IDF1$\uparrow$}
		& \textbf{HOTA$\uparrow$}
		& \textbf{IDSW$\downarrow$} \\
		\midrule
		\multirow{2}{*}{AIRMOT}
		& Online    & \textbf{46.10} & 36.19 & 31.94 & 26.15 & 400 \\
		& GT        & \textbf{46.10} & \textbf{36.27} & \textbf{31.97} & \textbf{26.20} & \textbf{398} \\
		\midrule
		\multirow{2}{*}{UAVSwarm}
		& Online    & \textbf{89.30} & 84.71 & 90.19 & 69.16 & 54 \\
		& GT        & \textbf{89.30} & \textbf{84.86} & \textbf{90.22} & \textbf{69.17} & \textbf{53} \\
		\bottomrule
	\end{tabular*}
\end{table}

As shown in Table~\ref{tab:gt_online_tracklet_gap}, the Online setting achieves performance very close to the GT setting on both datasets. The MOTA/IDF1/HOTA gaps are only 0.08/0.03/0.05 percentage points on AIRMOT and 0.15/0.03/0.01 percentage points on UAVSwarm, with only 2 and 1 additional ID switches, respectively, while AP50 remains unchanged. These results indicate that PSMP can use online tracklets for motion-prior construction with only marginal degradation in the final tracking performance.

\subsubsection{Effect of Probabilistic Prior Representation}

To examine whether explicitly retaining target-wise probabilistic components provides additional benefit, we compare the default moment-matched compact Gaussian prior with an explicit Gaussian mixture representation. All training and inference settings remain unchanged, and the two variants differ only in the probabilistic representation used for motion-prior-guided feature compensation. The default PSMP aggregates the target-wise distributions into a single Gaussian through moment matching, whereas the comparison retains them as an explicit mixture:
\begin{equation}
	\label{eq:gmm_prior}
	\begin{aligned}
		p_{\mathrm{mix}}(\delta)
		&= \sum_{i=1}^{n_T}\alpha_i
		\mathcal{N}(\delta;\hat{\mu}_i,\hat{{\Sigma}}_i),
		\mathcal{M}_{t,\mathrm{mix}}
		=\{\hat{\mathbf{v}}_{sw,t}, p_{\mathrm{mix}}(\delta)\}
	\end{aligned}
\end{equation}
where $\hat{\mu}_i$ and $\hat{{\Sigma}}_i$ are the predicted mean and covariance of the spatial offset distribution for the $i$-th reliable tracklet, and $\alpha_i$ is the corresponding normalized aggregation weight. The normalized mixture distribution is then used within the same EEC Activation pipeline.

\begin{table}[!t]
	\centering
	\setlength{\tabcolsep}{1.8pt}
	\renewcommand{\arraystretch}{1.05}
	\caption{Comparison between the compact Gaussian prior and the explicit Gaussian mixture prior in PSMP. ``Compact" denotes the compact Gaussian prior, and ``Mixture" denotes the Gaussian mixture prior.}
	\label{tab:compact_vs_gmm}
	\begin{tabular*}{\columnwidth}{@{\extracolsep{\fill}}llcccc@{}}
		\toprule
		\textbf{Dataset} 
		& \textbf{Prior}
		& \textbf{MOTA$\uparrow$}
		& \textbf{IDF1$\uparrow$}
		& \textbf{HOTA$\uparrow$}
		& \textbf{IDSW$\downarrow$} \\
		\midrule
		\multirow{2}{*}{AIRMOT}
		& Compact
		& \textbf{36.19} & \textbf{31.94} & \textbf{26.15} & \textbf{400} \\
		& Mixture
		& 36.08 & 31.40 & 25.49 & 421 \\
		\midrule
		\multirow{2}{*}{UAVSwarm}
		& Compact
		& \textbf{84.71} & \textbf{90.19} & \textbf{69.16} & \textbf{54} \\
		& Mixture
		& 84.28 & 89.42 & 68.83 & 66 \\
		\bottomrule
	\end{tabular*}
\end{table}

As shown in Table~\ref{tab:compact_vs_gmm}, the compact Gaussian prior achieves slightly but consistently better performance than the explicit Gaussian mixture prior on both datasets. On AIRMOT, it improves MOTA, IDF1, and HOTA by 0.11, 0.54, and 0.66 percentage points, respectively, and reduces IDSW by 21. On UAVSwarm, the corresponding improvements are 0.43, 0.77, and 0.33 percentage points, with 12 fewer ID switches. These results indicate that explicitly preserving the target-wise Gaussian components does not provide additional benefit for the proposed feature compensation under the evaluated settings.

\subsection{Ablation Study of EEC Activation}
To further analyze the effectiveness of EEC Activation, we conduct ablation studies from four aspects: the Local EEC score, the local window size used for residual statistics, the channel-adaptive gating strategy, and qualitative visualization of the intermediate feature compensation process. 

\subsubsection{Effectiveness of the Local EEC Score}
To evaluate the effectiveness of the Local EEC score, we compare different residual-statistics-based activation strategies within the same HOMATracker-based compensation framework. All variants use the same PSMP module and downstream detection-and-tracking framework, while differing in how the motion-prior-projected historical features are evaluated and weighted before fusion with the current-frame features.

Specifically, we compare four variants: (1) Direct fusion, which directly fuses motion-prior-projected historical features with current-frame features without EEC Activation; (2) Entropy-only, which employs only the residual entropy term in EEC Activation; (3) Energy-only, which uses only the feature residual energy term in EEC Activation; (4) Local EEC, which jointly incorporates residual energy and residual entropy. For the entropy-only, energy-only, and Local EEC variants, the same channel-adaptive gating strategy is used to generate the soft consistency mask. The weighted historical features are then fused with the current-frame features and fed into the downstream detection and association modules. The results are reported in Table~\ref{tab:eec_score_ablation}.

\begin{table*}[!t]
	\centering
	\setlength{\tabcolsep}{2.0pt}
	\renewcommand{\arraystretch}{1.0}
	\caption{Ablation of the Local EEC score on AIRMOT and UAVSwarm. E, S, and C.A. denote residual energy, residual entropy, and channel-adaptive gating, respectively.}
	\label{tab:eec_score_ablation}
	\begin{tabular*}{\textwidth}{@{\extracolsep{\fill}}lccc*{5}{c}*{5}{c}@{}}
		\toprule
		\multirow{2}{*}{\textbf{Variant}}
		& \multirow{2}{*}{\textbf{E}}
		& \multirow{2}{*}{\textbf{S}}
		& \multirow{2}{*}{\textbf{C.A.}}
		& \multicolumn{5}{c}{\textbf{AIRMOT}}
		& \multicolumn{5}{c}{\textbf{UAVSwarm}} \\
		\cmidrule(lr){5-9} \cmidrule(lr){10-14}
		&
		&
		&
		& \textbf{AP50$\uparrow$}
		& \textbf{MOTA$\uparrow$}
		& \textbf{IDF1$\uparrow$}
		& \textbf{HOTA$\uparrow$}
		& \textbf{IDSW$\downarrow$}
		& \textbf{AP50$\uparrow$}
		& \textbf{MOTA$\uparrow$}
		& \textbf{IDF1$\uparrow$}
		& \textbf{HOTA$\uparrow$}
		& \textbf{IDSW$\downarrow$} \\
		\midrule
		Direct Fusion
		& $\times$ & $\times$ & $\times$
		& 42.60 & 33.15 & 30.48 & 25.14 & 470
		& 85.80 & 82.49 & 88.96 & 67.79 & 69 \\
		
		Entropy-only
		& $\times$ & \checkmark & \checkmark
		& 44.90 & 33.78 & 31.03 & 25.48 & 447
		& 86.70 & 83.57 & 89.05 & 67.65 & 68 \\
		
		Energy-only
		& \checkmark & $\times$ & \checkmark
		& 45.40 & 35.60 & 31.50 & 25.62 & 428
		& 87.60 & 83.87 & 89.08 & 68.21 & 67 \\
		
		Local EEC
		& \checkmark & \checkmark & \checkmark
		& \textbf{46.10} & \textbf{36.19} & \textbf{31.94} & \textbf{26.15} & \textbf{400}
		& \textbf{89.30} & \textbf{84.71} & \textbf{90.19} & \textbf{69.16} & \textbf{54} \\
		\bottomrule
	\end{tabular*}
\end{table*}

As shown in Table~\ref{tab:eec_score_ablation}, the variants with residual-statistics-based activation generally outperform direct fusion on both datasets. This indicates that motion-prior-guided projection alone is insufficient, because the projected historical features may also carry background interference and spatially mismatched responses caused by motion uncertainty. Compared with direct fusion, both the Entropy-only and Energy-only variants improve most detection and tracking metrics, demonstrating the benefit of explicit cross-frame consistency evaluation. Among them, the energy-only variant is generally more effective than the entropy-only variant, indicating that residual magnitude provides a stronger cue than spatial-dispersion information alone for evaluating cross-frame feature consistency.

The full Local EEC score achieves the best overall performance on both datasets. On AIRMOT, compared with the energy-only variant, Local EEC improves AP50, MOTA, IDF1 and HOTA by 0.70, 0.59, 0.44 and 0.53 percentage points, respectively, while reducing IDSW by 28. On UAVSwarm, the corresponding improvements are 1.70, 0.84, 1.11, 0.95 percentage points, with 13 fewer ID switches. This demonstrates that the joint use of residual energy and residual entropy enables more effective motion-prior-conditioned feature consistency evaluation for selective feature compensation.

\subsubsection{Effect of Local Window Size in Local EEC}
To analyze the effect of different local support regions in Local EEC, we conduct an ablation study on the neighborhood window used to compute local residual statistics. In this experiment, the PSMP module, EEC Activation and the downstream detection-and-tracking framework are kept unchanged, and only the neighborhood size used in Local EEC is varied. Specifically, we evaluate four window sizes, $1\times 1$, $3\times 3$, $5\times 5$, and $7\times 7$ on AIRMOT and UAVSwarm. The results are reported in Table~\ref{tab:local_window_size}. 

On AIRMOT, performance improves from $1\times1$ to $3\times3$ and then decreases with larger windows, with $3\times3$ achieving the best overall results. Compared with the $5\times 5$ setting, it improves AP50, MOTA, IDF1, and HOTA by 1.10, 0.37, 0.82 and 0.42 percentage points, respectively, while reducing IDSW by 20. This suggests that, on AIRMOT, where UAV targets are relatively small and feature responses are weak, the $1\times1$ window provides insufficient spatial support, whereas larger windows may introduce more background interference into the residual statistics. Therefore, the $3\times3$ window achieves a better balance between spatial support and interference suppression.
In contrast, UAVSwarm favors a larger spatial support, with $7\times7$ yielding the best performance despite minor fluctuations at intermediate window sizes. Compared with the $3\times 3$ setting, the $7\times 7$ window improves AP50, MOTA, IDF1, and HOTA by 0.90, 0.28, 0.70, and 0.51 percentage points, respectively, and reduces IDSW by 12. This suggests that the relatively larger target-response regions in UAVSwarm benefit from a wider local window, which captures more complete spatial information for consistency estimation.
Overall, these results indicate that the optimal local window size depends on the spatial extent of target responses: overly small windows may provide insufficient spatial support, whereas overly large windows may introduce excessive background responses.

\begin{table}[!t]
	\centering
	\setlength{\tabcolsep}{2.2pt}
	\renewcommand{\arraystretch}{1.0}
	\caption{Effect of local window size in Local EEC on AIRMOT and UAVSwarm.}
	\label{tab:local_window_size}
	\begin{tabular*}{\columnwidth}{@{\extracolsep{\fill}}lccccc@{}}
		\toprule
		\textbf{Window} 
		& \textbf{AP50$\uparrow$} 
		& \textbf{MOTA$\uparrow$} 
		& \textbf{IDF1$\uparrow$} 
		& \textbf{HOTA$\uparrow$} 
		& \textbf{IDSW$\downarrow$} \\
		\midrule
		\multicolumn{6}{@{}l}{\textbf{AIRMOT}} \\
		\midrule
		$1\times1$ & 43.50 & 35.49 & 30.62 & 25.35 & 423 \\
		$3\times3$ & \textbf{46.10} & \textbf{36.19} & \textbf{31.94} & \textbf{26.15} & \textbf{400} \\
		$5\times5$ & 45.00 & 35.82 & 31.12 & 25.73 & 420 \\
		$7\times7$ & 44.20 & 35.74 & 30.58 & 25.40 & 418 \\
		\midrule
		\multicolumn{6}{@{}l}{\textbf{UAVSwarm}} \\
		\midrule
		$1\times1$ & 87.40 & 83.58 & 89.42 & 68.41 & 73 \\
		$3\times3$ & 88.40 & 84.43 & 89.49 & 68.65 & 66 \\
		$5\times5$ & 88.30 & 84.31 & 89.23 & 68.54 & 67 \\
		$7\times7$ & \textbf{89.30} & \textbf{84.71} & \textbf{90.19} & \textbf{69.16} & \textbf{54} \\
		\bottomrule
	\end{tabular*}
\end{table}

\subsubsection{Effect of Channel-Adaptive Gating in EEC Activation}
To analyze the effect of channel-adaptive gating in EEC Activation, we compare two gating strategies while keeping the PSMP module, the local window size, the feature compensation framework, and the downstream detection and tracking framework unchanged. Shared gating uses a common activation threshold and scaling factor for all channels, whereas channel-adaptive gating learns channel-specific activation parameters. The results are reported in Table~\ref{tab:gating_ablation}.

As shown in Table~\ref{tab:gating_ablation}, channel-adaptive gating consistently outperforms shared gating on both AIRMOT and UAVSwarm. On AIRMOT, it improves AP50, MOTA, IDF1, and HOTA by 0.80, 0.50, 1.27, and 0.87 percentage points, respectively, while reducing IDSW by 34. On UAVSwarm, it improves AP50, MOTA, IDF1, and HOTA by 0.80, 0.24, 0.46, and 0.22 percentage points, respectively, and reduces IDSW by 19. These results suggest that residual-consistency responses vary across feature channels, making shared activation parameters less effective in capturing channel-wise differences. Channel-specific parameters allow EEC Activation to adapt to these channel-wise variations, thereby preserving motion-consistent historical responses while suppressing interfering responses more selectively.

\begin{table}[!t]
	\centering
	\setlength{\tabcolsep}{1.8pt}
	\renewcommand{\arraystretch}{1.0}
	\caption{Effect of channel-adaptive gating in EEC Activation on AIRMOT and UAVSwarm. ``Adaptive" denotes channel-adaptive gating.}
	\label{tab:gating_ablation}
	\begin{tabular*}{\columnwidth}{@{\extracolsep{\fill}}llccccc@{}}
		\toprule
		\textbf{Dataset} 
		& \textbf{Gating}
		& \textbf{AP50$\uparrow$}
		& \textbf{MOTA$\uparrow$}
		& \textbf{IDF1$\uparrow$}
		& \textbf{HOTA$\uparrow$}
		& \textbf{IDSW$\downarrow$} \\
		\midrule
		\multirow{2}{*}{AIRMOT}
		& Shared   & 45.30 & 35.69 & 30.67 & 25.28 & 434 \\
		& Adaptive & \textbf{46.10} & \textbf{36.19} & \textbf{31.94} & \textbf{26.15} & \textbf{400} \\
		\midrule
		\multirow{2}{*}{UAVSwarm}
		& Shared   & 88.50 & 84.47 & 89.73 & 68.94 & 73 \\
		& Adaptive & \textbf{89.30} & \textbf{84.71} & \textbf{90.19} & \textbf{69.16} & \textbf{54} \\
		\bottomrule
	\end{tabular*}
\end{table}

\subsubsection{Qualitative Analysis of EEC Activation}
\begin{figure*}[!t]
	\centering
	\includegraphics[width=\textwidth]{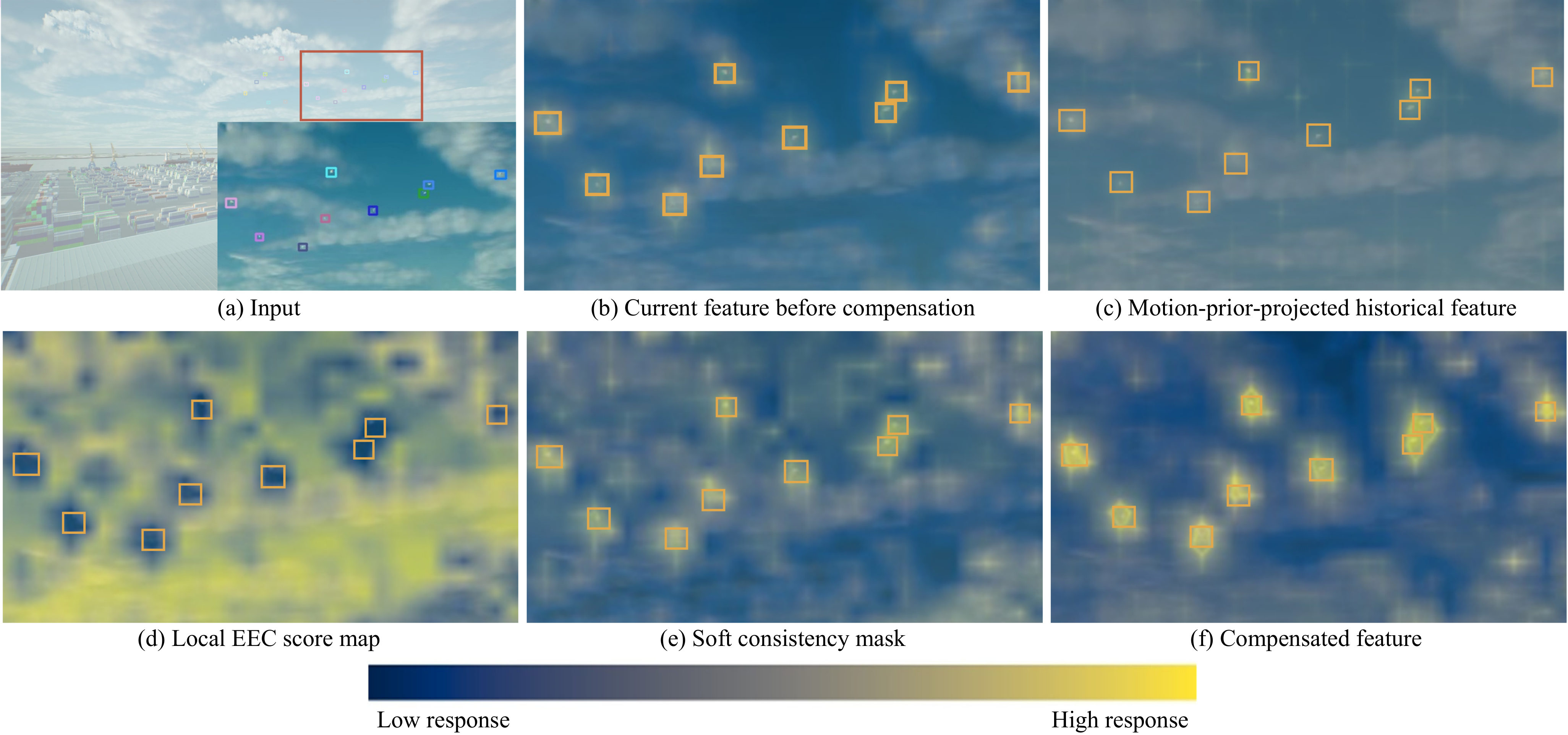}
	\caption{Visualization of the intermediate feature compensation process of EEC Activation on AIRMOT. The subfigures show (a) the input image patch, (b) the current feature before compensation, (c) the motion-prior-projected historical feature, (d) the Local EEC score map, (e) the channel-wise soft consistency mask, and (f) the final compensated feature map. The color bar indicates the response magnitude from low to high.}
	\label{fig_6}
\end{figure*}

To provide qualitative evidence for EEC Activation, we visualize the intermediate feature compensation process on a representative frame from the AIRMOT-02 sequence, as shown in Fig.~\ref{fig_6}. The figure presents the following stages: the input image patch, the current-frame feature before compensation, the motion-prior-projected historical feature, the Local EEC score map, the channel-wise soft consistency mask, and the final compensated feature map. These intermediate results illustrate how EEC Activation performs feature compensation through motion-prior-guided projection, Local EEC-based consistency evaluation, soft mask construction, and selective weighting followed by feature fusion.

Before feature compensation, the target response in the current frame is weak and partially obscured by background responses. After projecting the historical feature toward the current frame under the probabilistic swarm motion prior, the projected feature retains a clearer target response but also carries noticeable background interference. This indicates that motion-prior-guided projection alone is insufficient. The Local EEC score map further shows that the target response exhibits a relatively low score, whereas many interfering responses have higher values. This is consistent with the intended design of Local EEC: motion-consistent cross-frame target responses tend to produce lower residual energy and entropy, while less-consistent interfering responses generally exhibit larger residual magnitudes or higher spatial dispersion. Consequently, Local EEC provides a discriminative feature-space measure for separating motion-consistent potential target responses from interfering historical responses.

Based on this score map, the soft consistency mask converts low-score, high-consistency regions into larger activation weights for feature compensation. After applying the soft mask weighting and feature fusion, motion-consistent potential UAV target responses are strengthened and background interference is reduced in the compensated current-frame feature. These results show that EEC Activation does not simply propagate historical features. Instead, it uses the Local EEC score map to selectively incorporate motion-prior-projected historical information, thereby strengthening weak target feature responses and reducing background interference.

\subsection{Runtime Efficiency and Embedded Deployment}
For airborne optical UAV swarm tracking, online inference efficiency and embedded deployability are important system-level considerations. Therefore, in addition to tracking accuracy, we further evaluate a lightweight implementation of EECTracker on an NVIDIA Jetson Orin NX platform to assess its embedded deployment performance.

Considering the limited computational resources of embedded devices, we implement a lightweight variant, denoted as EECTracker-Light. It adopts HOMATracker-Light as the base framework, using YOLOX-S for detection together with a lightweight appearance branch for appearance embedding. PSMP and EEC Activation are then integrated into this lightweight tracking pipeline. We select the representative TBD method ByteTrack as the comparison method due to its efficiency and deployment flexibility. ByteTrack is also configured with the same YOLOX-S detector and denoted as ByteTrack-Light. The neural-network inference components of both methods are accelerated with TensorRT on Jetson Orin NX.

\begin{table}[!t]
	\centering
	\setlength{\tabcolsep}{3.0pt}
	\renewcommand{\arraystretch}{1.12}
	\caption{Embedded deployment performance on Jetson Orin NX.}
	\label{tab:nx_deployment}
	\begin{tabular}{llcccc}
		\toprule
		\textbf{Method} & \textbf{Dataset}
		& \textbf{MOTA$\uparrow$}
		& \textbf{IDF1$\uparrow$}
		& \textbf{HOTA$\uparrow$}
		& \textbf{FPS$\uparrow$} \\
		\midrule
		\multirow{2}{*}{ByteTrack-Light}
		& AIRMOT   & 23.46 & 24.58 & 22.86 & \textbf{24.6} \\
		& UAVSwarm & 63.71 & 75.13 & 59.60 & \textbf{23.9} \\
		\midrule
		\multirow{2}{*}{EECTracker-Light}
		& AIRMOT   & \textbf{31.68} & \textbf{30.03} & \textbf{24.79} & 21.0 \\
		& UAVSwarm & \textbf{78.73} & \textbf{86.35} & \textbf{65.60} & 21.3 \\
		\bottomrule
	\end{tabular}
\end{table}

As shown in Table~\ref{tab:nx_deployment}, EECTracker-Light achieves 31.68\% MOTA, 30.03\% IDF1, and 24.79\% HOTA at 21.0 FPS on AIRMOT, improving MOTA, IDF1, and HOTA over ByteTrack-Light by 8.22, 5.45, and 1.93 percentage points, respectively. On UAVSwarm, it achieves 78.73\% MOTA, 86.35\% IDF1, and 65.60\% HOTA at 21.3 FPS, with corresponding improvements of 15.02, 11.22, and 6.00 percentage points over ByteTrack-Light. Although EECTracker-Light incurs a modest reduction in inference speed, it maintains an end-to-end rate above 20 FPS on both datasets while providing substantial tracking-performance gains. These results indicate that EECTracker can be adapted to a lightweight online tracking pipeline while retaining both competitive tracking accuracy and practical inference efficiency, supporting its deployment in embedded airborne optical UAV swarm tracking systems.

\section{Conclusion}
This paper addresses stable online UAV swarm tracking under challenging airborne optical imaging conditions, where small target scales and complex background interference can weaken target feature responses and disrupt temporal observations. We presented EECTracker, a swarm-motion-prior-guided joint detection and tracking framework that exploits swarm-level historical motion information to compensate potential UAV target responses. Within EECTracker, the proposed feature compensation strategy constructs a probabilistic swarm motion prior from available reliable historical tracklets by modeling the shared short-term motion tendency of the swarm and its uncertainty. The resulting prior characterizes statistically reachable image-plane regions and provides spatial guidance for historical feature projection. EEC Activation further evaluates motion-prior-conditioned cross-frame feature consistency and selectively incorporates projected historical responses over potential target regions through channel-adaptive soft gating. The compensated feature representation is then used for detection and association, while updated tracklets support subsequent motion-prior construction in an online closed-loop manner.

Extensive experiments on AIRMOT and UAVSwarm demonstrate that the proposed swarm-motion-prior-guided feature compensation strategy provides robust swarm-level motion guidance for downstream compensation, while EEC Activation effectively modulates the influence of projected historical responses according to motion-prior-conditioned feature consistency. Building on these components, EECTracker achieves superior overall tracking performance compared with state-of-the-art methods. The Jetson Orin NX experiments further indicate the potential of EECTracker for embedded deployment on airborne platforms. A current limitation is that the benefit of probabilistic swarm motion guidance decreases when reliable historical tracklets become highly sparse. Future work will focus on more robust swarm motion modeling under severely limited historical observations.

\section*{Acknowledgments}
This study was co-supported by the National Key Research and Development Program of China (2023YFC3341100), National Natural Science Foundation of China (No. 52672527).
\bibliographystyle{IEEEtran}
\bibliography{IEEEabrv,reference}
%
\vspace{6pt}

\bf{}\vspace{-33pt}

\begin{IEEEbiography}[{\includegraphics[width=1in,height=1.25in,clip,keepaspectratio]{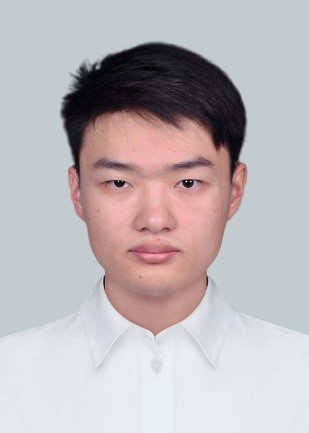}}]{Zhaochen Chu}
	received the B.E. degree in Flight Vehicle Design and Engineering from Beijing Institute of Technology, Beijing, China, in 2021. He is currently pursuing the Ph.D. degree with the China-UAE Belt and Road Joint Laboratory on Intelligent Unmanned Systems with the School of Aerospace Engineering, Beijing Institute of Technology. His research interests include small target UAV detection and swarm UAV tracking based on visual imagery.
\end{IEEEbiography}

\begin{IEEEbiography}
	[{\includegraphics[width=1in,height=1.25in,clip,keepaspectratio]{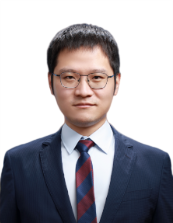}}]{Tao Song}
	received the B.E. degree in Guidance, Navigation and Control, and Ph.D. degree in Aircraft Design from Beijing Institute of Technology, Beijing, China, in 2008 and 2014, respectively. He is currently an Associate Professor with the School of Aerospace Engineering, Beijing Institute of Technology. His research interests include UAV swarm systems, intelligent aircraft modeling, and guidance and control.
\end{IEEEbiography}

\begin{IEEEbiography}
	[{\includegraphics[width=1in,height=1.25in,clip,keepaspectratio]{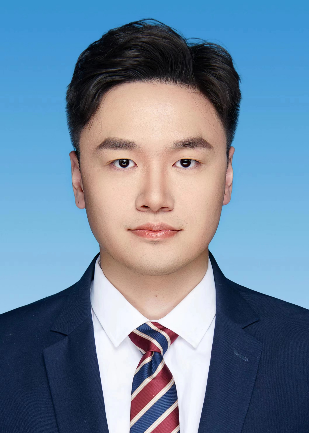}}]{Ren Jin}
	received the M.E. degree in Computer Application Technology from Hefei University of Technology, Hefei, China, in 2016, and the Ph.D. degree in Aerospace Science and Technology from Beijing Institute of Technology, Beijing, China, in 2020. He is currently a Tenure-Track Assistant Professor with the School of Aerospace Engineering, Beijing Institute of Technology. His research interests include onboard visual object detection, recognition and tracking, and UAV visual navigation.
\end{IEEEbiography}

\begin{IEEEbiography}
	[{\includegraphics[width=1in,height=1.25in,clip,keepaspectratio]{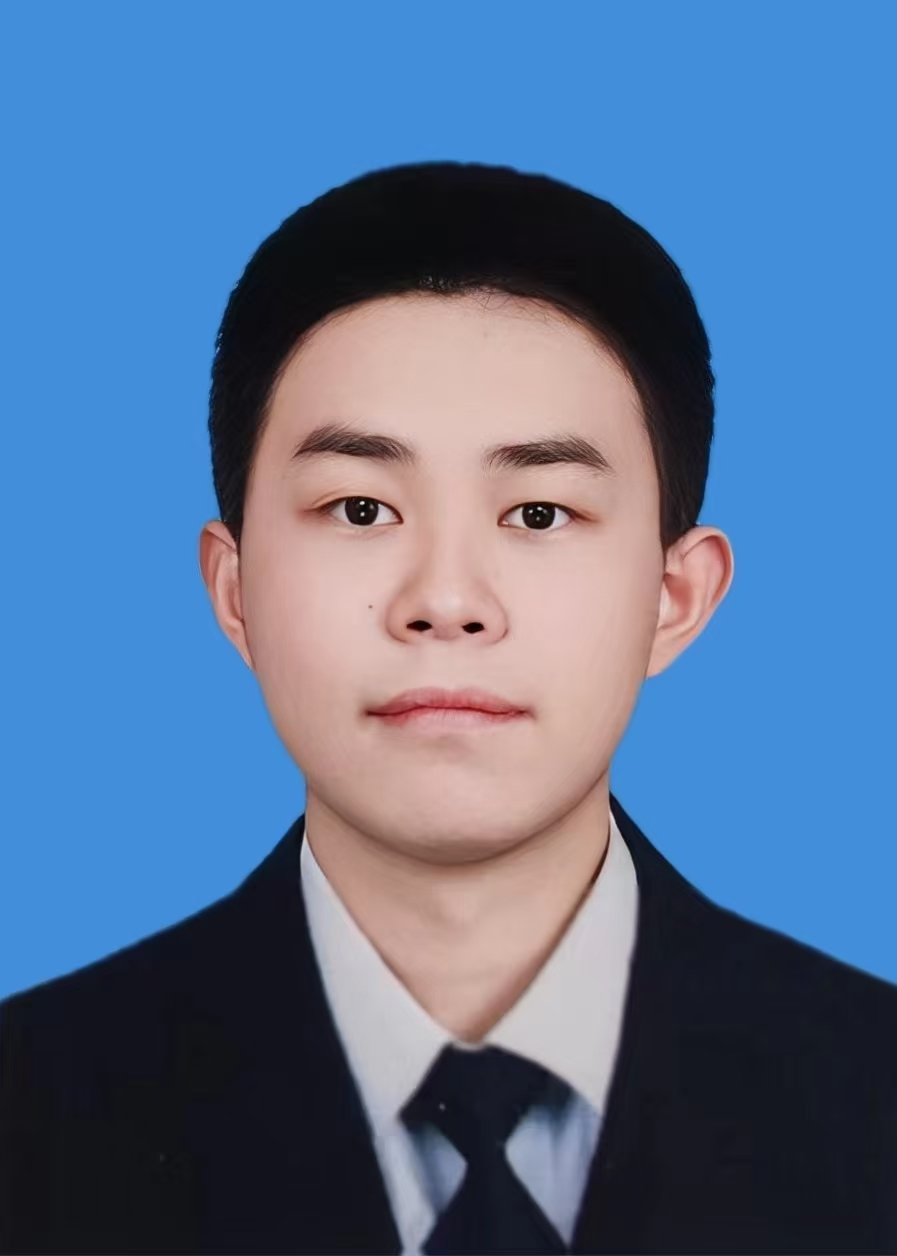}}]{Mingdong Jia}
	received the B.E. degree in Information Security from Tianjin University of Technology, Tianjin, China, in 2024. He is currently pursuing the M.S. degree with Beijing Institute of Technology, Beijing, China. His research interests include UAV vision, object detection and multi-object tracking.
\end{IEEEbiography}

\begin{IEEEbiography}
	[{\includegraphics[width=1in,height=1.25in,clip,keepaspectratio]{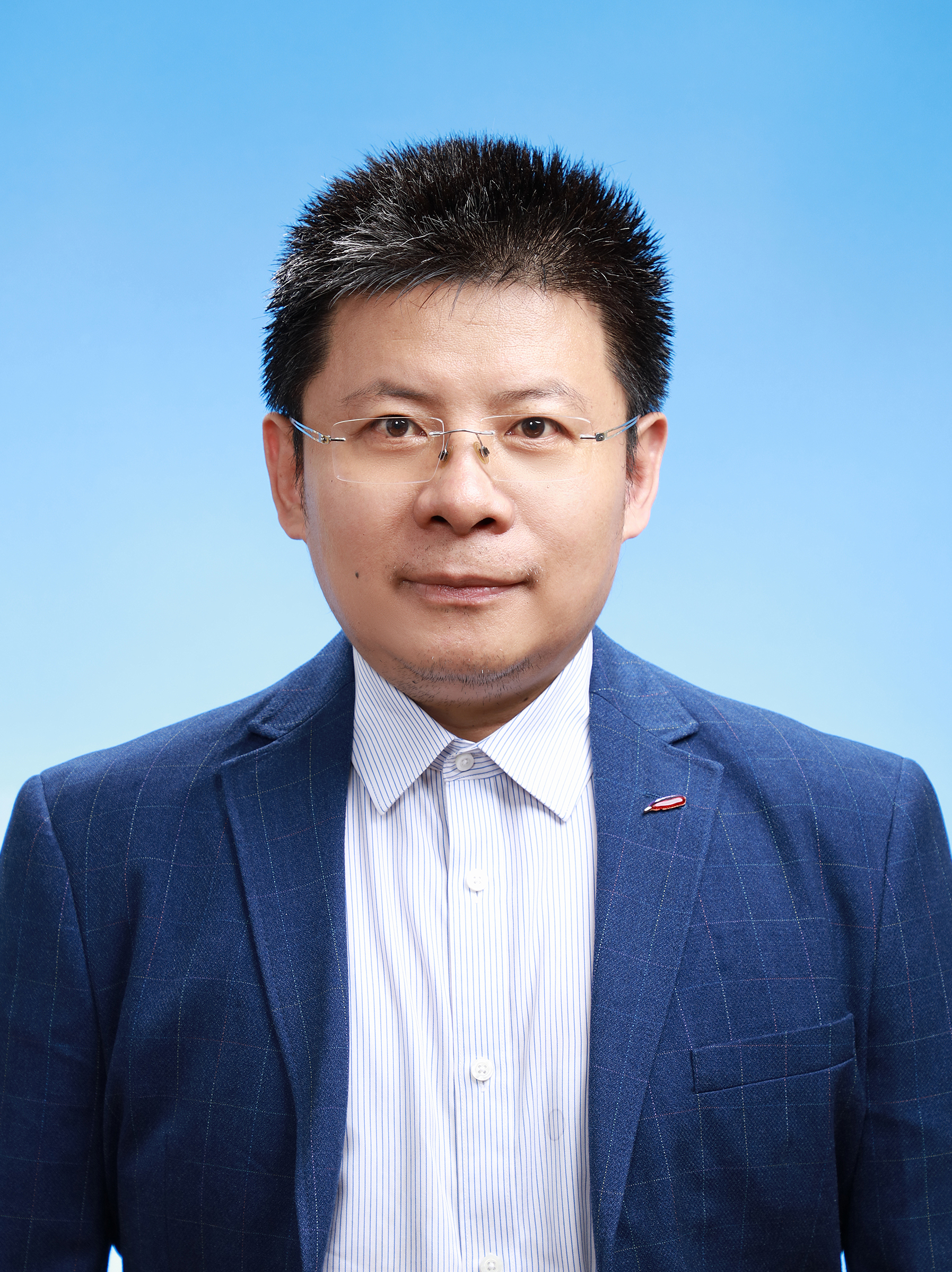}}]{Defu Lin}
	received the M.E. and Ph.D. degrees in Aircraft Design from Beijing Institute of Technology, Beijing, China, in 1999 and 2005, respectively. He is currently a Professor with the School of Aerospace Engineering, Beijing Institute of Technology. He directs the Beijing Key Laboratory for UAV Autonomous Control and the China-UAE Belt and Road Joint Laboratory on Intelligent Unmanned Systems. His research interests include aircraft system design, flight vehicle guidance, and control technologies.
\end{IEEEbiography}

\vspace{4pt}

\vfill

\end{document}